\documentclass[12pt]{iopart}

\usepackage{bm}
\usepackage{algpseudocode}
\usepackage{algorithmicx}
\usepackage[section]{algorithm}
\usepackage{hyperref}
\hypersetup{
    colorlinks=true,
    linkcolor=blue,
    filecolor=magenta,      
    urlcolor=cyan,
    pdftitle={Overleaf Example},
    pdfpagemode=FullScreen,
    }
\usepackage{multicol}
\usepackage{epsfig}

\usepackage{iopams}  
\begin{document}

\title{On tuning a mean-field model for semi-supervised classification}

\author{Emílio Bergamim \& Fabricio Breve}

\address{Avenida 24A, 1515, Rio Claro, State of São Paulo, Brazil}
\ead{emiliobergjr@gmail.com}
\vspace{10pt}
\begin{indented}
\item[]February 2022
\end{indented}

\begin{abstract}
Semi-supervised learning (SSL) has become an interesting research area due to its capacity for learning in scenarios where both labeled and unlabeled data are available. In this work, we focus on the task of transduction - when the objective is to label all data presented to the learner - with a mean-field approximation to the Potts model. Aiming at this particular task we study how classification results depend on $\beta$ and find that the optimal phase depends highly on the amount of labeled data available. In the same study, we also observe that more stable classifications regarding small fluctuations in $\beta$ are related to configurations of high probability and propose a tuning approach based on such observation. This method relies on a novel parameter $\gamma$ and we then evaluate two different values of the said quantity in comparison with classical methods in the field. This evaluation is conducted by changing the amount of labeled data available and the number of nearest neighbors in the similarity graph. Empirical results show that the tuning method is effective and allows NMF to outperform other approaches in datasets with fewer classes. In addition, one of the chosen values for $\gamma$ also leads to results that are more resilient to changes in the number of neighbors, which might be of interest to practitioners in the field of SSL.
\end{abstract}

%\maketitle

\section{Introduction}

Semi-supervised learning (SSL) has become a major area of interest in machine learning due to its capacity for learning in the presence of labeled and unlabeled data. Conceptually, SSL is mainly seen as a midpoint between the major areas of \textit{unsupervised} and \textit{supervised} learning \cite{chapelle2006,van2020}.
\par 
In unsupervised tasks, the main goal is to determine the intrinsic structure of a dataset $D$, such as estimating a density function over it or partitioning it into subsets whose inner elements carry some form of similarity (also known as \textit{clustering}). Supervised learning, on the other hand, aims to learn a function $y(x)$ from a sample of pairs $D = \{(x_{i},y_{i})\}_{i=1}^{N}$, like in the case of regression and classification \cite{alpaydin2004}.
\par 
The applicability of these two paradigms is usually defined according to the amount of labeled data available to the learner: unsupervised learning is suited for tasks where no labeled data is available \textit{a priori}, while supervised learning is used when $D$ is completely labeled \cite{mohri2018}.
Such an extreme separation would lead to asking what to do when only a fraction of the elements of $D$ are labeled. This is the scenario where \textit{semi-supervised classification} takes place \cite{van2020}.
\par 
Under partially labeled data, one may be interested in learning a rule to label unseen data or in labeling all available data. The first situation is known as \textit{inductive}, while the second is called \textit{transductive} semi-supervised classification \cite{van2020}.
\par 
Algorithms for transduction generally rely on representing $D$ through a graph for which each node represents an instance and weighted edges among these represent the similarity of a pair of elements of $D$ \cite{van2020,chong2020}. Transductive algorithms have found recent applications in many areas such as image segmentation \cite{breve2019}, online tracking \cite{zhang2016}, text recognition in images \cite{liu2020} and video object segmentation \cite{zhang2020}.

\subsection{Contributions}

Our work revolves around semi-supervised transduction with the Potts model \cite{wu1982potts}. Previous works with similar models have pointed out the problem of tuning the $\beta$ parameter as the main obstacle to the employment of these models both in clustering \cite{ott2004} and transduction \cite{wang2007,li2016} tasks. 
\par 
We then study how classification results depend on $\beta$ by analyzing results in a range of the said parameter. In this investigation, we find that the optimal phase is highly dependent on the amount of labeled data available and that a more stable region for choosing $\beta$ is associated with the probability of the most probable configuration of the system, $\Gamma$. We then construct an approximation for $\Gamma$ and propose tuning $\beta$ by choosing a target value for $\Gamma$.
\par 
Next, we compare the proposed approach with classical SSL approaches using two different targets for $\Gamma$. To do so, we investigate the behavior of said algorithms in different constructions of the similarity graphs and different rates of labeled data. We find our tuning approach is effective and can outperform existing approaches on some datasets, especially when the number of classes is small. 
\par 
Also, one of the tuning methods introduced showed to be particularly resilient to changes in the number of neighbors in the graph. This property is particularly useful, since finding an optimal topology for the similarity graph is a difficult task.

\subsection{Organization}

The work follows with a brief exposure of transductive semi-supervised classification's particularities and the classical algorithms for this task. We then discuss the Potts model and the Naive Mean-Field Approximation that allows for the approximate calculation of statistical properties and connects this approach with other transductive ones through a propagation algorithm. Next, the experimental setup is introduced and followed by two different sections on experimental results, the first regarding the dependency and tuning of $\beta$ and the second consisting of comparisons with classical algorithms. 

\section{Transductive semi-supervised classification}

\par 
Let $D = \{(\mathbf{x}_{i},y_{i})\}_{i=1}^{N}$ be a dataset for which $\mathbf{x}_{i} \in \mathbb{R}^{n}$ denotes the attributes of the $i$-th sample and $y_{i} \in \{1,...,q\}$ its label. In a SSL setup there is a subset $D_{l}\subset D$ for which the labels are kwown in advance and another subset $D_{u} = D\setminus D_{l}$ of unlabeled samples. 
\par 
For transductive methods, the goal is to label all elements of $D$. To do so, one first constructs a \textit{weighted similarity graph} $G_{D}$ for which each node is mapped to a sample of $D$ and each edge describes the similarity among neighboring samples. Next, one applies an inference algorithm to the graph to label each node. These are the two main steps of semi-supervised transduction \cite{van2020,chong2020}, which we will now discuss in more depth.

\subsection{Transductive inference in graphs}
\par
The goal of a transductive algorithm is to use the information available in $D_{l}$ to label points in $D_{u}$ \cite{chong2020}. To do so, one needs a similarity matrix $\mathbf{W}$ for which entries $W_{i,j}\in [0,1]$ describe the similarities between the $i$-th and $j$-th samples in $D$. The nonzero entries in $\mathbf{W}$ correspond to the (weighted) edges of $G_{D}$.
\par 
Once such objects are available to the learner, the inference phase consists of a propagation algorithm where the \textit{a priori} knowledge in $D_{l}$ is propagated over $G_{D}$ (and weighted by $\mathbf{W}$) to all samples in $D$. In this context, two algorithms are currently established as the standard approaches: \textit{Gaussian Random Fields} (GRF) and \textit{Local and Global Consistency}(LGC) \cite{van2020,chong2020}, the first being regarded as the state of the art by some authors \cite{karasuyama2017,ma2020}. 
\par 
If there are $q$ possible labels in a set, let $\phi_{i}(s_{i})$ be the probability that the $i$-th sample in $D$ has label $s_{i}\in \{1,...,q\}$. GRF \cite{zhu2002,zhu2005} is an approach that minimizes the objetive function
\begin{equation}
H_{GRF}(\boldsymbol{\phi}) = -\frac{1}{2}\sum_{i,j}W_{i,j}\sum_{s}[\phi_{i}(s)-\phi_{j}(s)]^{2}
\end{equation}
under the constraint that instances in $D_{l}$ have their probabilities campled to
\begin{equation}
	\phi_{i}(s_{i}) = \Bigg\{
	\begin{array}{l@{}l}
		1, \textrm{ if } s_{i}=y_{i} \\
		0 \textrm{ otherwise.}
	\end{array}
\end{equation}
\par 
LGC \cite{zhou2004}, on the other hand, was proposed as a way to overcome limitations of GRF regarding noisy labels or irregular graph structures and is also widely used \cite{van2020,chong2020}. In this case, one does not work with probabilities in the strict sense, but with a set of vectors $\{\mathbf{f}_{i}\}_{i=1}^{N}$ in $\pmb{R}^{q}$ that minimizes 
\begin{equation}
H_{LGC}(\mathbf{f}) = \sum_{i,j}W_{i,j}||\mathbf{f}_{i}-\mathbf{f}_{j}||^{2} + \frac{1-\alpha}{\alpha}\sum_{i}||\mathbf{f}_{i}-\boldsymbol{\theta}_{i}||^{2},
\end{equation}
where $\boldsymbol{\theta}_{i} \in \pmb{R}^{q}$ is defined by
\begin{equation}\label{eq:fields}
	\theta_{i,s} = \Bigg\{
	\begin{array}{l@{}l}
		1, \textrm{ if } (x_{i},y_{i})\in D_{l} \textrm{ and } s=y_{i},\\
		0, \textrm{ otherwise}
	\end{array}
\end{equation}
and $\alpha \in (0,1)$ is the parameter of such model\cite{zhou2004}.
\par 
An exact approach to minimize the objetives in (1) and (4) requires the inversion of $\mathbf{W}$, for which complexity is $O(N^{3})$ \cite{van2020,zhou2004}. However, both methods allow for an iterative solution with complexity $O(N^{2})$. These are presented in algorithms \autoref{grf} and \autoref{lgc}. After convergence of such procedures the points are labeled as
\begin{equation}
y_{i} = \textrm{argmax}_{s_{i}}{\phi_{i}(s_{i})} \quad \textrm{and} \quad y_{i} = \textrm{argmax}_{s_{i}}{f_{i,s_{i}}}
\end{equation}
for GRF and LGC, respectively.

	\begin{algorithm}
		\caption{Iterative GRF \label{grf}}
		\begin{algorithmic}
		\State \textbf{Input}: $\boldsymbol{\phi}^{(0)}$, $\mathbf{W}$, $D_{l}$, $\epsilon$, $t_{max}$
		\State $t \gets 0$, $\delta \gets \epsilon$
		\While{$t<t_{max}$ and $\delta \geq \epsilon$}
			\State $\delta \gets 0$
			\For{$i=1,...,N$}
				\If{$(x_{i},y_{i}) \notin D_{l}$}
				\For {$s=1,...,q$}
					\State $\phi_{i}^{(t+1)}(s) \gets \frac{\sum_{j\neq i}W_{i,j}\phi_{j}^{(t)}(s)}{\sum_{l}\sum_{j\neq i}W_{i,j}\phi_{j}^{(t)}(l)}$
				\EndFor
				\EndIf
				\State $\delta \gets \max\{\delta,\max_{s}|\phi_{i}^{(t+1)}(s)-\phi_{i}^{(t)}(s)|\}$
			\EndFor
			\State $t \gets t+1$
		\EndWhile
		\State \textbf{Output: } $\boldsymbol{\phi}^{(t)}$
		\end{algorithmic}
	\end{algorithm}
	
	\begin{algorithm}
		\caption{Iterative LGC \label{lgc}}
		\begin{algorithmic}
		\State \textbf{Input}: $\mathbf{f}^{(0)}$, $\mathbf{W}$, $\boldsymbol{\theta}$, $\alpha$, $\epsilon$, $t_{max}$
		\State $t \gets 0$, $\delta \gets \epsilon$
		\While{$t<t_{max}$ and $\delta \geq \epsilon$}
			\State $\delta \gets 0$
			\For{$i=1,...,N$}
				\For {$s=1,...,q$}
					\State $f_{i,s}^{(t+1)} \gets \alpha\sum_{j\neq i}W_{i,j}f_{j,s}^{(t)}+(1-\alpha)\theta_{i,s}$
				\EndFor
				\State $\delta \gets \max\{\delta,\max_{s}|f_{i,s}^{(t+1)}-f_{i,s}^{(t)}|\}$
			\EndFor
			\State $t \gets t+1$
		\EndWhile
		\State \textbf{Output: } $\mathbf{f}^{(t)}$
		\end{algorithmic}
	\end{algorithm}

\subsection{Similarity construction}
	\label{subsec22}
Both algorithms presented before are examples of propagation dynamics on graphs. In fact, by examining these methods, one can see that the weighting process becomes irrelevant if $G_{D}$ describes the relations among elements of $D$ in such a way that edges only connect points that belong in the same class. Weighting becomes a necessity since one usually does not know in advance a sufficient topology of $G_{D}$ for class detection.
\par 
In this work a $k$ nearest neighbors (kNN) approach is used for construction of $\mathbf{W}$: the nonzero entries of the $i$-th row of $\mathbf{W}$ are the $k$ nearest neighbors of $x_{i}$ according to a distance function $d(\cdot,\cdot)$.
\par 
A highly popular method for weighting edges uses a gaussian radial basis function \cite{van2020} 
\begin{equation}\label{eq:rbf}
	W_{i,j} = \exp\bigg\{ \frac{d(\mathbf{x}_{i},\mathbf{x}_{j})^{2}}{2\sigma^{2}} \bigg\}
\end{equation}
and the parameter $\sigma$ can be tuned as \cite{jebara2009}
\begin{equation}\label{eq:sigma}
\sigma = \frac{1}{3N}\sum_{i}d_{i,k(i)},
\end{equation}
where $k(i)$ denotes the $k-$th nearest neighbor of the $i-$th element of $D$. A previous study using several weighting schemes reported this approach as the most effective when combined with different inference algorithms \cite{deSousa2013}.
\par 
It is also a common place in the literature to use a sparse construction of $\mathbf{W}$ \cite{deSousa2013,van2020,chong2020} where most of its entries are set to zero, leaving only the most significant entries in order to increase contrast among different classes. Choosing how sparse one wants $\mathbf{W}$ to be is the problem of choosing a particular $k$ to the problem.
\par 
In \cite{deSousa2013} it was verified that a robust way of obtaining a sparse similarity is by setting the non-zero entries of $\mathbf{W}$ to initially be the $k-$th nearest neighbor of each sample in $D$ and tune $\sigma$ as in \autoref{eq:sigma}. This is then followed by a symmetrization and sparsification procedure from which a novel similarity is obtained via
\begin{equation}
	W_{i,j} \gets \min\{W_{i,j},W_{j,i}\}. 
\end{equation} 
Then, $\mathbf{W}$ is normalized by setting the sum of its rows to 1:
	\begin{equation}\label{eq:norm}
		W_{i,j} \gets \frac{W_{i,j}}{\sum_{l\neq i}W_{i,l}}.
	\end{equation}
	\par 
	In the present work we will focus on the above similarity construction with the distance function $d(\cdot,\cdot)$ being the euclidean (L2) norm and evaluate algorithms for different values of $k$. For further reading on the problem of constructing $\mathbf{W}$ we refer the reader to \cite{deSousa2013,subramanya2014,van2020,chong2020}.

\section{Potts model and naive mean fields}\label{sec:potts}
The main hypothesis of semi-supervised learning is that elements of $D$
with the same label belong to the same cluster - i.e., the classes of the dataset are the labeling of different clusters \cite{chapelle2006,van2020}.
\par 
Under this consideration, transduction can be viewed as the optimization of a cost function $H_{C}(\mathbf{s})$ over a set of labels $\mathbf{s} = (s_{1},...,s_{N})$ under the restriction that known labels are constants. In fact, the case
\begin{equation}\label{eq:potts_C}
H_{C} = -\sum_{i<j}W_{i,j}\delta(s_{i},s_{j}),
\end{equation}
where $\delta(\cdot,\cdot)$ is the Kronecker Delta, is known to be equivalent to the GRF algorithm when variables corresponding to labeled data are held fixed \cite{tibely2008}.

\par 
\autoref{eq:potts_C} is known as the Potts model \cite{wu1982potts} without an external field. In other works, this model has been used in the presence of an external field as defined for the LGC model in \autoref{eq:fields} \cite{wang2007,li2016}. In this case, the cost function is of the form
\begin{equation}\label{eq:potts}
H(\mathbf{s}) = -\sum_{i,j}W_{i,j}\delta(s_{i},s_{j})-\sum_{i,s}\theta_{i,s}\delta(s_{i},s).
\end{equation}
\par 
Also, the latter approach is probabilistic since labes are attributed according to the distribution
\begin{equation}\label{eq:potts_p}
\Psi(\mathbf{s}) = \frac{1}{Z_{H}}\exp\{-\beta H(\mathbf{s})\}, 
\end{equation}
where $Z_{H}$ is a normalization constant known as the partition function, for which calculation imposes a problem, as its exact evaluation demands one to sum over all $q^{N}$ possible configurations of $\mathbf{s}$.
\par 
A possible overcome of this problem is the usage of Monte Carlo methods, as was done both in clustering (corresponding to the zero-field form of \autoref{eq:potts}) and transduction \cite{blatt1996,shental2005} applications. This, however, has the problem of relying on random number generation, which renders some unpredictability to the behavior of the algorithms.
\par 
A deterministic approach to the problem is the use of mean-field methods, more recently studied both in clustering \cite{shi2018} and transduction \cite{wang2007,li2016}. In this case, one seeks to approximate \autoref{eq:potts_p} by a more tractable distribution, with the easier way of doing so being known as the Naive Mean Field (NMF) method \cite{opper2001,wang2007}.
\par 
This approach consists of finding a distribution $\Phi(\mathbf{s})$ that minimizes the Kullback-Leibler divergence
\begin{equation}\label{eq:dkl}
	D_{KL}(\Phi||\Psi) = \sum_{\mathbf{s}}\Phi(\mathbf{s})\ln \bigg\{ \frac{\Phi(\mathbf{s})}{\Psi(\mathbf{s})}\bigg\}
\end{equation}
under the constraint that $\Phi$ is a product distribution: 
\begin{equation}
	\Phi(\mathbf{s}) = \prod_{i}\phi_{i}(s_{i}).
\end{equation}

	\par 
	Under this construction $\Phi$ is a distribution that considers $\mathbf{s}$ as a vector of independent variables. Therefore, $\phi_{i}(s_{i})$ is the marginal distribution of $s_{i}$. Minimization of \autoref{eq:dkl} with respect to such marginals results in a set of non-linear coupled equations
	\begin{equation}\label{eq:nmf_shape}
		\phi_{i}(s_{i}) = \frac{1}{Z_{i}}\exp\{h_{i}(s_{i})\}, \textrm{ with } Z_{i} = \sum_{s_{i}}\exp\{h_{i}(s_{i})\}
	\end{equation}
	and 
	\begin{equation} \label{eq:nmf_h}
		h_{i}(s_{i}) = \beta \bigg(\theta_{i,s_{i}}+\sum_{j\neq i} W_{i,j}\phi_{j}(s_{i})\bigg)
	\end{equation}		 
that can be solved iteratively with complexity $O(N^{2})$ as shown in Algorithm \autoref{nmf}. After this procedure, one labels instances of $D$ in a similar way as is done for GRF and LGC:
\begin{equation}
	y_{i} = \textrm{argmax}_{s_{i}}\phi_{i}(s_{i}).
\end{equation}

\begin{algorithm}
		\caption{Iterative NMF \label{nmf}}
		\begin{algorithmic}
		\State \textbf{Input}: $\Phi^{(0)}$, $\beta$, $\mathbf{W}$, $\boldsymbol{\theta}$, $\epsilon$, $t_{max}$
		\State Initialize $\mathbf{h}^{(0)} = \mathbf{0}$
		\State $t \gets 0$, $\delta \gets \epsilon$
		\While{$t<t_{max}$ and $\delta \geq \epsilon$}
			\State $\delta \gets 0$
			\For{$i=1,...,N$}
				\For {$s=1,...,q$}
					\State $h_{i}(s)^{(t+1)} \gets \beta\bigg(\theta_{i,s_{i}}+\sum_{j\neq i}W_{i,j}\phi_{j,s}^{(t)}\bigg)$
				\EndFor
				\State Calculate $\phi_{i}^{(t+1)}$ using \autoref{eq:nmf_shape}.
				\State $\delta \gets \max\{\delta,\max_{s}|h_{i}(s)^{(t+1)}-h_{i}(s)^{(t)}|\}$
			\EndFor
			\State $t \gets t+1$
		\EndWhile
		\State \textbf{Output: } $\Phi^{(t)}$
		\end{algorithmic}
	\end{algorithm}
	
	\subsection{Previous works on tuning $\beta$}
	
	We will work with the NMF approximation to calculate the statistical properties of the Potts model, the remaining problem is to understand how the classification results will be affected by $\beta$. 
	\par 
	Earlier works on clustering (corresponding to the zero-field form \autoref{eq:potts}) advocate for the existence of a range for $\beta$ where results are optimal in some sense \cite{blatt1996}. It is known, however, that in such a range different clustering structures exist and the idea of optimality can only emerge by analyzing and combining these different possible clusterings \cite{ott2004}. 
	\par 
	The application of these ideas to classification is followed immediately by fixing variables associated with elements of $D_{l}$ to known labels and applying the same tuning procedure to estimate $\beta$. It was also observed that increasing the size of $D_{l}$ also increases the optimal range for $\beta$ \cite{shental2005}.
	\par 
	A major drawback of the above works is the necessity for evaluating the statistical properties of the model for different values of $\beta$ in order to determine the optimal range. Despite using Monte Carlo methods, the usage of mean-field methods would not offer a significant improvement to this, especially in the semi-supervised classification problem, since GRF is a well-established non-parametric approach. 
	\par 
	Regarding mean-field methods, more recent work was done with the Ising model \cite{wang2007,li2016} using a similar similarity construction as the one outlined in the previous section. However, the authors in these works highlighted that the need for efficient tuning of $\beta$ is the main obstacle to practical usage, since results on accuracy point that this approach reaches state-of-the-art performance \cite{wang2007,li2016}.
	\par 
	It is also noteworthy that earlier work on SSL with the Potts model uses $\beta-$independent fields set to be infinitely strong \cite{shental2005}, while most recent approaches \cite{wang2007,li2016} have focused on $\beta-$ dependent fields, as is the case of the model in \autoref{eq:potts_p}. Both approaches do so in order to keep labeled data fixed to their known labels, but in different ways. The first approach freezes variables to their known labels. The second allows marginals of variables with non-zero fields to change with $\beta$, but not their labels. To better understand this case, we draw attention to the normalization of similarities (\autoref{eq:norm}) and the fact that marginals are upper-bounded by unity to obtain the following inequalities for the NMF equations (\autoref{eq:nmf_h})	
	\begin{equation}
	\beta\theta_{i,s_{i}} \leq h_{i}(s_{i}) \leq \beta(\theta_{i,s_{i}}+1).
	\end{equation}
	Then, the definition of fields (\autoref{eq:fields}) implies that if the $m-$th instance of $D$ is labeled as $y_{m}$ we have
	\begin{equation}
	h_{m}(s_{m})\leq h_{m}(y_{m}).
	\end{equation}
	We then note that equality in the above expression can only be achieved for $s_{m}\neq y_{m}$ and $\beta\neq 0$ if 
	\begin{equation}
		\sum_{j\neq m}W_{m,j}(\phi_{j}(s_{m})-\phi_{j}(y_{m})) = 1
	\end{equation}
	and normalization of similarities implies this can only be achieved if $\phi_{j}(s_{m})=1$ for all neighbors of $m$. This would imply marginals of neighbors of $m$ to be unnafected by $\phi_{m}$, which is true only if we have $W_{m,j}=0$ together with $W_{j,m}=0$ which is not impossible under the similarity construction discussed in \autoref{subsec22}. We then conclude that algorithm \autoref{nmf} is not able to change known labels in $D$.
	\par 
	Finally, we also highlight recent work on clustering using a Potts spin-glass together with mean-field methods \cite{shi2018}, which related optimal results to a phase transition. Our work, however, will focus on ferromagnetic interactions to follow along previous works on semi-supervised classification with mean-field methods that were proven effective in this task \cite{wang2007,li2016}.
	\par 
	Next, we discuss the experimental setup that will be used for the remainder of this paper in order to understand the $\beta$ dependency problem and compare NMF, LP and GRF.

\section{Experimental setup}
	
	As we aim to evaluate semi-supervised methods for different configurations of the problem, here we discuss the data that will be used for such evaluation. Particularly, three bidimensional artificially generated datasets will be used together with other six high-dimensional datasets available in the literature.
	\par
\begin{figure}[h]
	\centering
	\includegraphics[scale=0.35]{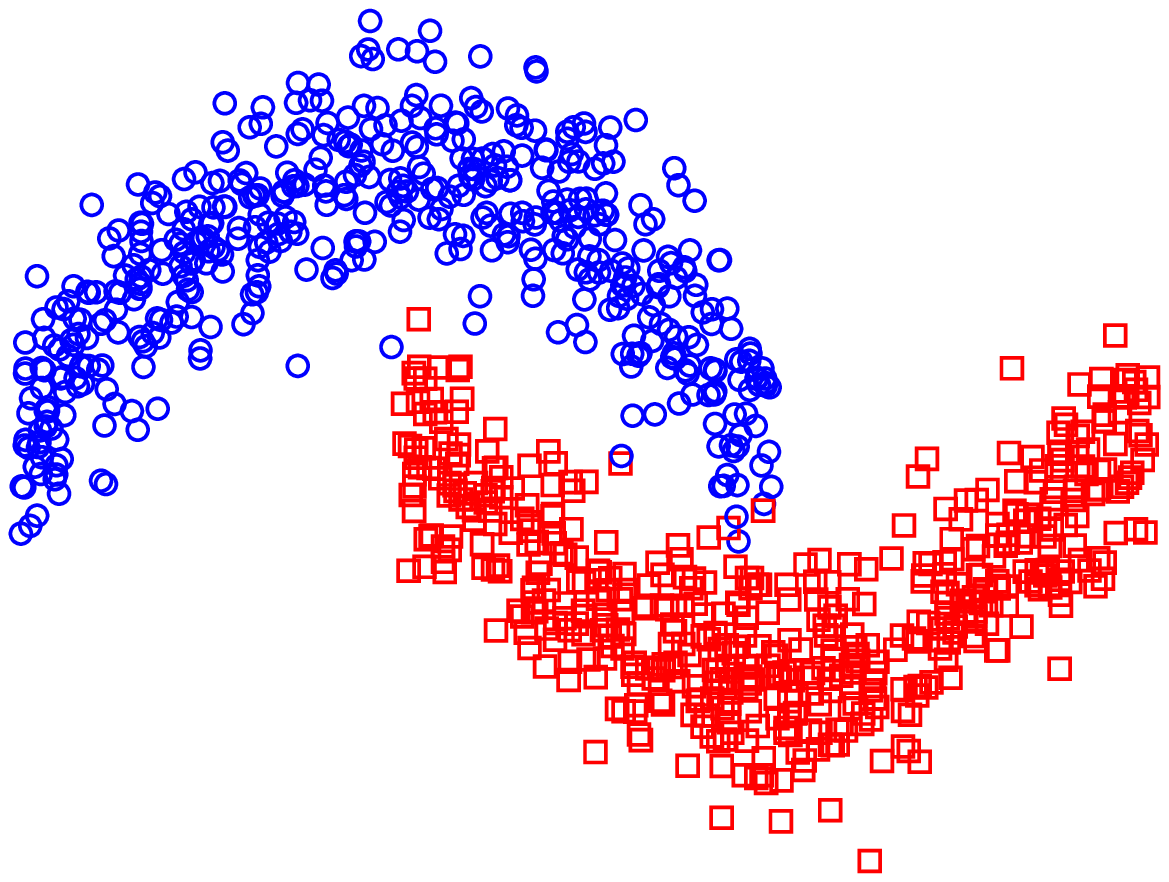}
	\includegraphics[scale=0.35]{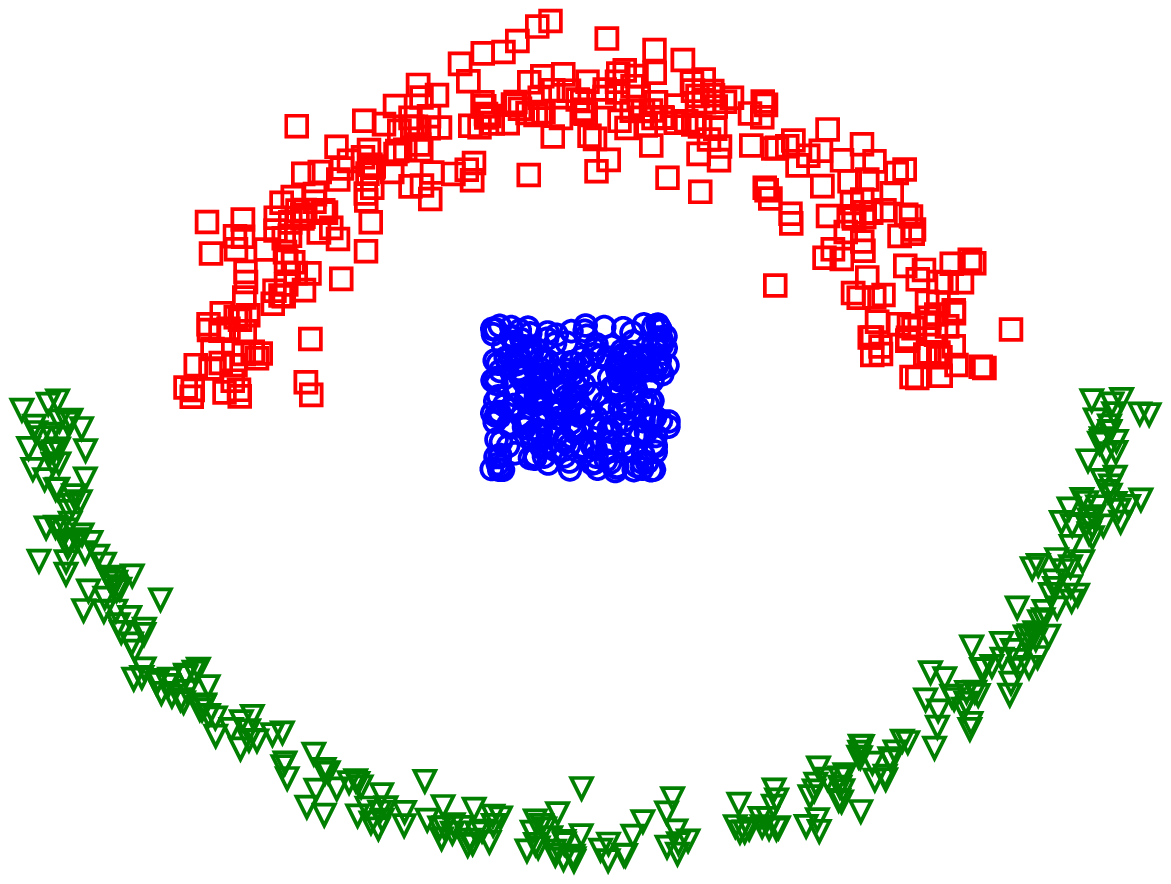}
	\includegraphics[scale=0.35]{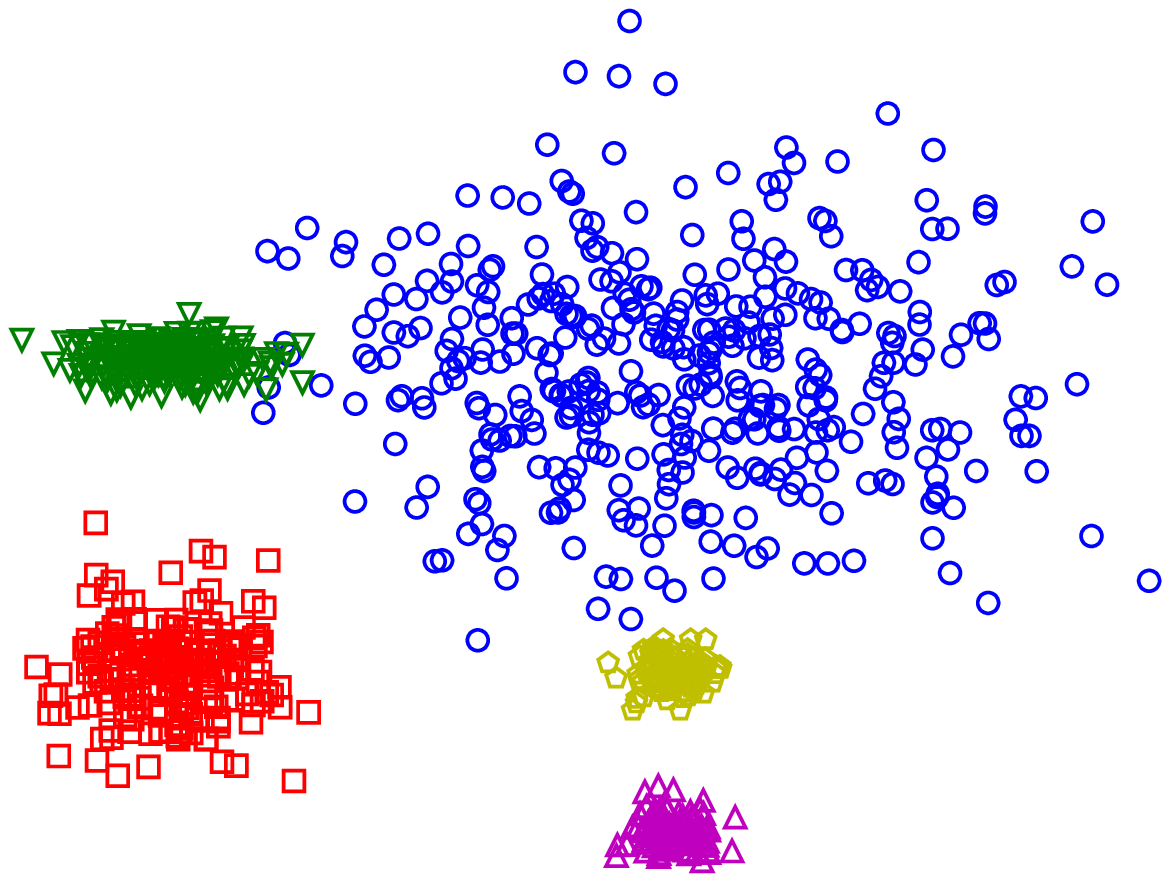}
	\caption{Images of the three bidimensional datasets used for benchmarking: two moons (top left), three clusters (top right) and five gaussians (bottom). Different symbols relate to different labels.\label{fig1}}
\end{figure}

 
	The bidimensional datasets (\autoref{fig1}) are \textbf{Two moons}, a set consisting of two balanced non-convex classes; \textbf{Three clusters}, a set of three unbalanced classes, with two of them being non-convex and \textbf{Five Gaussians} with different locations and standard deviations, as well as different proportions. The first and last datasets contain $1000$ elements, while the second contains $900$. 
	
	\par 
	The high-dimensional datasets are 
\begin{itemize}
	\item \textbf{Digit1} was introduced in \cite{chapelle2006} to evaluate semi-supervised algorithms. It has $N=1500$, dimension $d=241$ and was preprocessed by the original authors. It is a set of artificial images of the digit ``1'', divided in two classes \cite{chapelle2006}.
	\item \textbf{Twonorm} and \textbf{Ringnorm} are available in the DELVE repository \cite{delve}. These are sets of 7400 instances in a 20-dimensional space consisting of two gaussian distributions. In each dataset classes are differed by their means and covariance matrices.
	\item \textbf{Landsat} is a dataset of 6435 $3\times 3$ hyperspectral satellite images divided in $7$ classes. It was obtained from the UCI repository \cite{Dua2019}.
	\item \textbf{USPS} is a set of handwritten digits from ``0'' to ``9'', but with $9298$ images \cite{hull1994} and dimensions $16 \times 16$. 
	\item \textbf{Texture} is a set from the ELENA project consisting of $5500$ patterns describing $11$ different textures (the classes). Each instance is described by $40$ attributes estimated from fourth-order modified moments. As the original repository is missing, we refer the reader to the version in \cite{alcala2011}. 
\end{itemize}

	\par 
	We evaluate algorithms over different constructions of $D_{l}$. Letting $N_{l}$ be the number of elements in $D_{l}$ and $r_{l} = N_{l}/N$ the rate of labeled data presented to the learner, for each value in the grid $r_{l} \in [0.02:0.2:0.02]$ we randomly generate twenty different realizations of $D_{l}$ assuring that at least one instance of each class is presented. 	
	\par 
	Classification results are then evaluated according to the accuracy and adjusted mutual information (AMI, evaluated in \textit{nats}) \cite{vinh2010}. The second is a metric commonly used to evaluate different clusterings of a dataset. Due to the clustering hypothesis of semi-supervised learning, comparing the accuracy and AMI will allow us to have a better understanding of how clustering connects to classification in these models.
	\par 
To run experiments on the described datasets we implement the propagation algorithms \autoref{grf}, \autoref{lgc} and \autoref{nmf} in the C programming language \cite{ritchie1988} and then build an interface in Cython \cite{behnel2010} to make these functions callable via Python \cite{python3}.
\par 
The Python part of our experiments is mainly the handling of input and output for each dataset, while algorithms \autoref{grf}, \autoref{lgc} and \autoref{nmf} are executed in C using multithreaded parallelism via the OpenMP standard \cite{dagum1998}. All codes used for benchmarking are available at \href{https://github.com/boureau93/ssl-nmf}{https://github.com/boureau93/ssl-nmf}. Experiments were carried out on an Intel i5-1135G7 processor with 8GB RAM.

\section{Dependency of classifications on $\beta$}
	\label{sec:5}

	Our first experimental evaluation aims to study how classification results depend on $\beta$. As we wish to develop a tuning procedure for the said parameter, we must relate said results to a statistical property. In this work we will focus on the mode probability 
	\begin{equation}
		\Gamma = \prod_{i}\phi_{i}(y_{i})
	\end{equation}
	and how this quantity connects to the accuracy and AMI at different values of $\beta$.
	\par 
	Figures 2 through 6 show results for accuracy, AMI, execution time and $\Gamma$ as a function of $\beta$ for three different values of $r_{l}$ and a fixed topology of $G_{D}$ at $k=\log_{2} N$. In these experiments we also set $t_{max} = 10^{4}$ and $\epsilon=10^{-3}$ and change $\beta$ in the grid $10^{[-3:3:0.2]}$.
	\par 
	Experiments on the artificial bidimensional datasets (Figures 2-4) show that optimal classification results are associated with higher values of $\Gamma$. When $\Gamma = (\frac{1}{q})^{N}$ the model is in an equiprobable configuration, leading to the worst classification results since one cannot effectively distinguish different labels via their probabilities. The increase of $\beta$ then leads to the increase of $\Gamma$. We also observed that optimal results are usually associated with $\Gamma$ above the midpoint between equiprobable probabilities and maximum probability, i.e., $\Gamma \geq (\frac{1+q}{2q})^{N}$.
	\par 
	As our experiments were done using $k=\log_{2}N$ (Figures 2-6), we expect most edges to connect only nodes that belong to the same class, but some inter-class connectivity still exists. This is overcome by the increase in $\beta$, which increases correlations among variables and, together with the information given by labeled data increases contrast among classes by making wrongly connected nodes less relevant.
	\par 
	However, the existence of a range for $\beta$ where accuracy and AMI are stable at a maximum seems to be conditioned on the amount of labeled data available. On the two moons (\autoref{fig3}) dataset we see this behavior: for $r_{l}=0.02$ there is a peak of optimal classification that is followed by a slow decrease in accuracy and AMI, while for the higher values of $r_{l}$ the model reaches an optimal plateau of such quantities. 
	\par 	
	On the two high-dimensional real datasets evaluated (\autoref{fig5} and \autoref{fig6}) we see the absence of the optimal classification plateau for all evaluated values of $r_{l}$. The Landsat dataset shows a more stable optimal region than USPS, which has a more pronounced peak in accuracy and AMI.
	\par 
	This behavior is in line with previous studies \cite{shental2005} pointing that the addition of labeled data increases the range of $\beta$ containing optimal results. On top of that, our results show that the existence of an optimal plateau where classifications are not affected by changes in $\beta$ is conditioned on the size of $D_{l}$, while the sufficient amount for the said phenomenon to occur is dataset dependent. We can then conclude that increasing $r_{l}$ increases the stability of classification metrics once $\Gamma$ is big enough. As labeled data becomes scarcer, the difference between optimal results and results in the $\Gamma$ plateau becomes more evident.
	\par 
	It is also notable that at higher $\Gamma$ the execution time of the algorithm scatters to a plateau of higher execution that is up to three orders of magnitude slower than the low $\Gamma$ region. Together with the decrease in accuracy and AMI at the $\Gamma$ plateau shown by some datasets we conclude this is a region where elements at the boundary of a class can be difficult to label due to their high correlations to elements of different classes, demanding more iterations of algorithm \autoref{nmf}. This can be evidenced by looking at the three clusters dataset , in which classes are more separated (\autoref{fig1}) and the execution time does not present a plateau at higher $\Gamma$.
	\par 
	The above also indicates that better classifications are not strictly associated with a higher execution time of the algorithm. What is true, however, is that higher values of $\Gamma$ are associated with a decrease in computational performance that may or may not be in the form of a plateau. 
	\par 	
	In fact, for sufficiently high $r_{l}$, close to optimal values of accuracy and AMI can coexist with lower execution times as illustrated in the results for bidimensional datasets (Figures 2-4). Therefore, for an appropriate choice of $\beta$, increasing the amount of labeled data can lead to better computational performance. 
	\par 
	Still regarding execution time, at higher values of $\beta$ the algorithm can slow down up to three orders of magnitude when compared to the paramagnetic phase where $\Gamma \approx (\frac{1}{q})^{N}$.
	\par 
	We also highlight that, for higher values of $\beta$ the used hardware lacks numerical precision, and solutions of the NMF equation return NaN values, leaving us unable to calculate $\Gamma$. So, the decrease in accuracy, AMI, and execution at higher $\beta$ may not be associated with the model or the approximation, but due to an experimental limitation.

	\begin{figure}[h]
		\centering
		\includegraphics[scale=0.5]{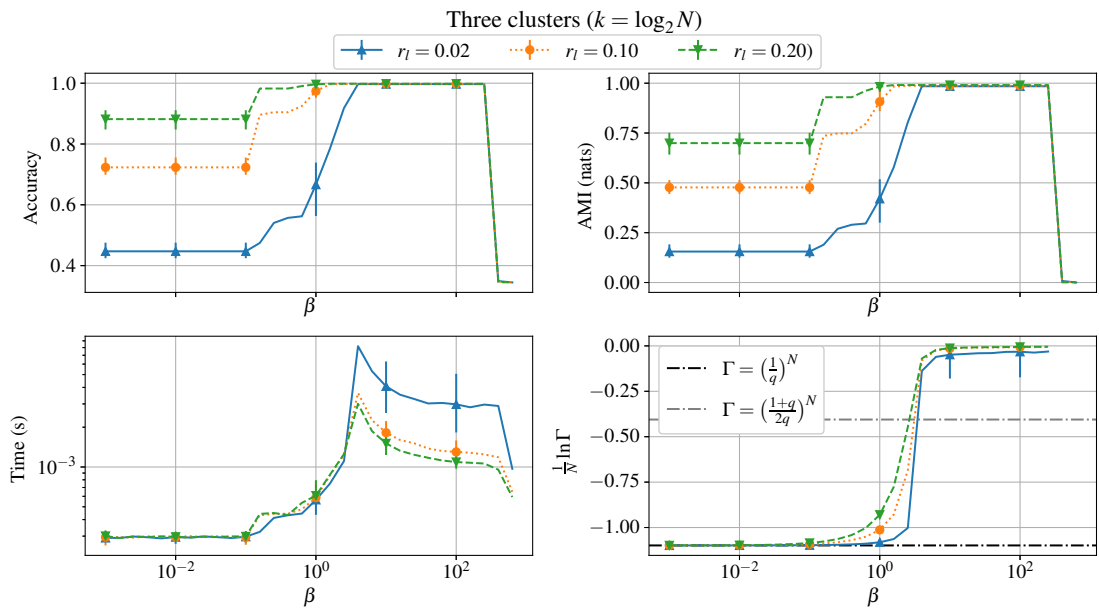}
		\caption{Results for the three clusters dataset as a function of $\beta$. Lines denote averages  and bars denote maximum and minimum over different realizations of $D_{l}$. \label{fig2}}
	\end{figure}
	
	\begin{figure}[h]
		\centering
		\includegraphics[scale=0.5]{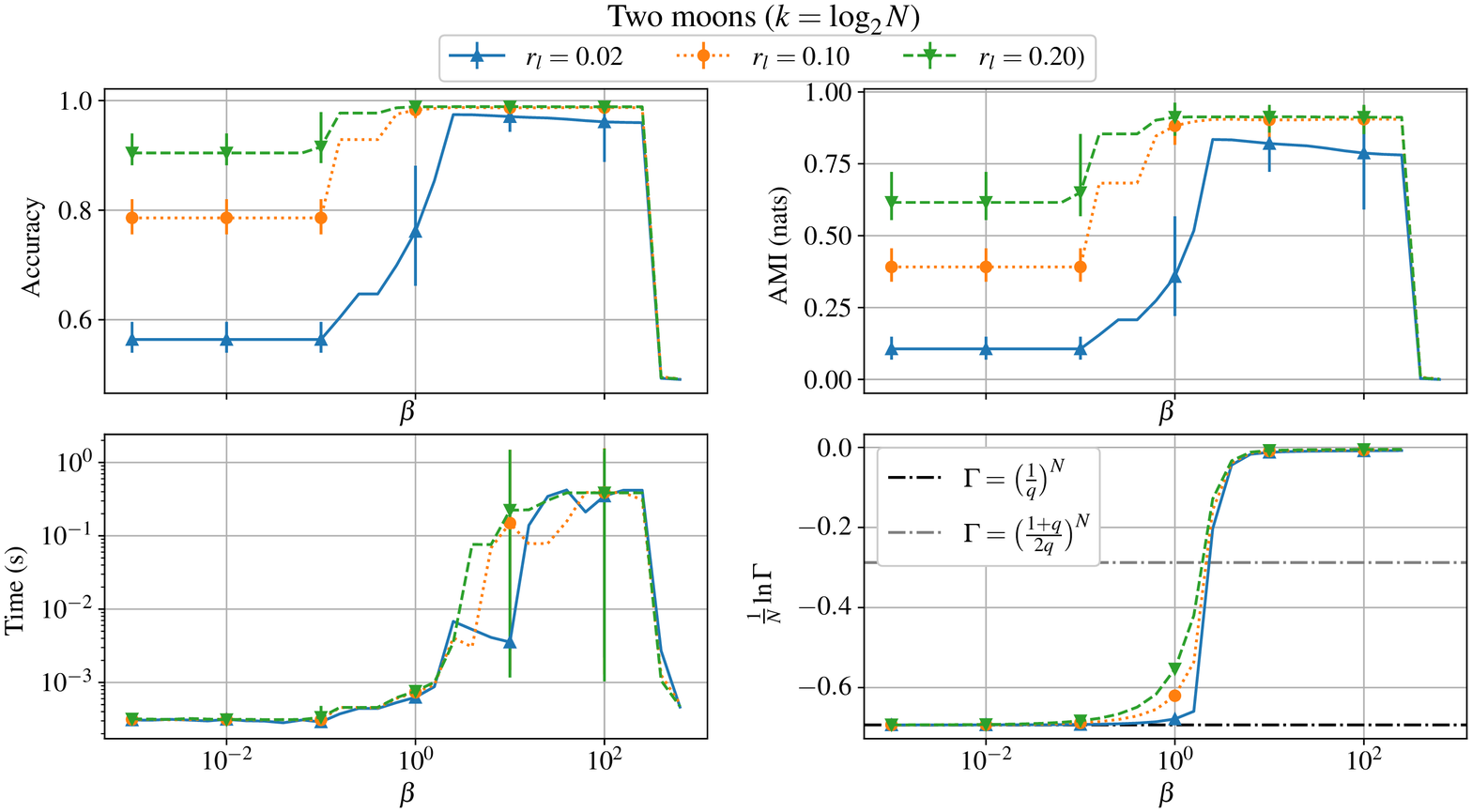}
		\caption{Results for the two moons dataset as a function of $\beta$. Lines denote averages  and bars denote maximum and minimum over different realizations of $D_{l}$. \label{fig3}}
	\end{figure}
	
	\begin{figure}[h]
		\centering
		\includegraphics[scale=0.5]{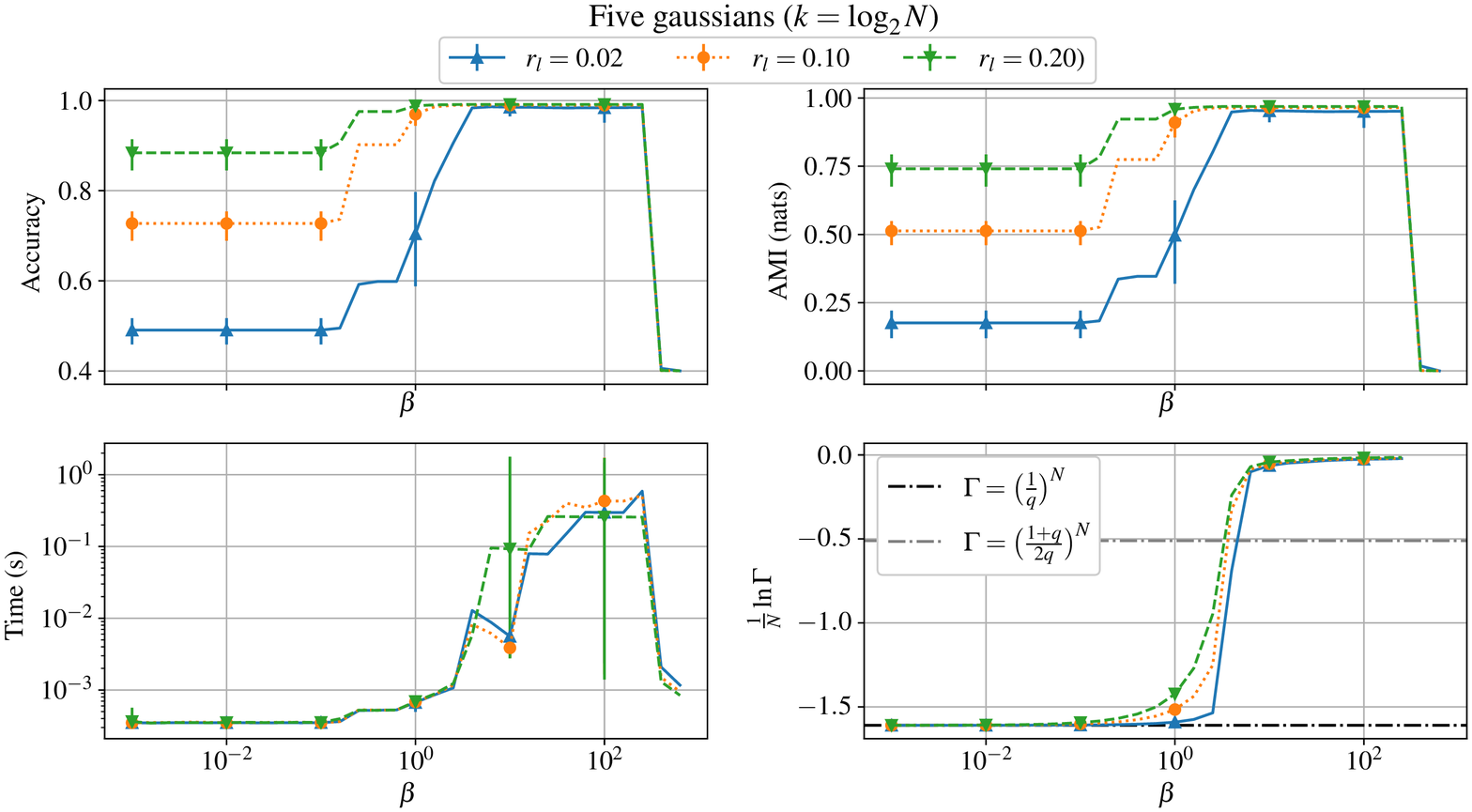}
		\caption{Results for the five gaussians dataset as a function of $\beta$. Lines denote averages  and bars denote maximum and minimum over different realizations of $D_{l}$. \label{fig4}}
	\end{figure}
	
	\begin{figure}[h]
		\centering
		\includegraphics[scale=0.5]{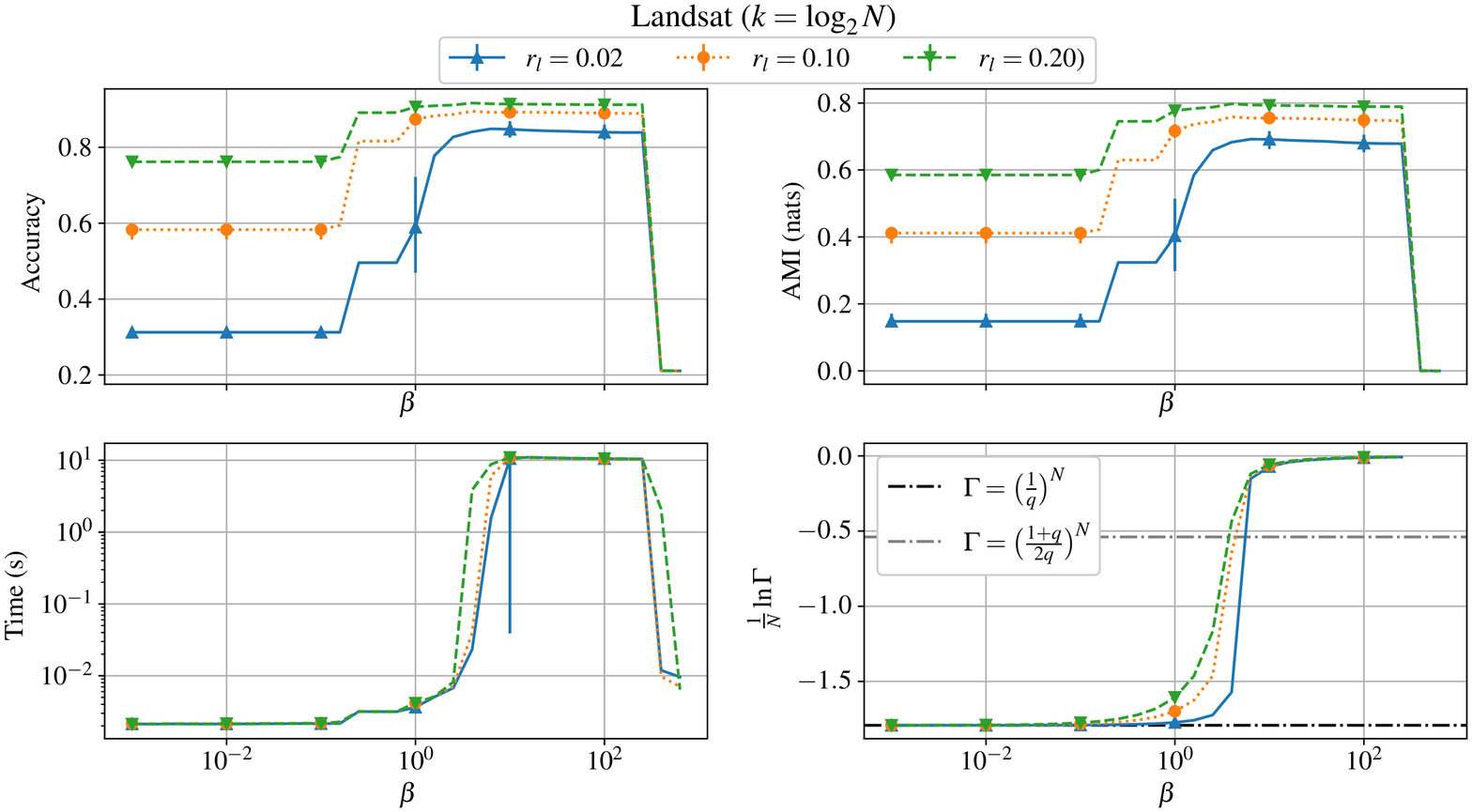}
		\caption{Results for the Phoneme dataset as a function of $\beta$. Lines denote averages  and bars denote maximum and minimum over different realizations of $D_{l}$. \label{fig5}}
	\end{figure}
	
	\begin{figure}[h]
		\centering
		\includegraphics[scale=0.5]{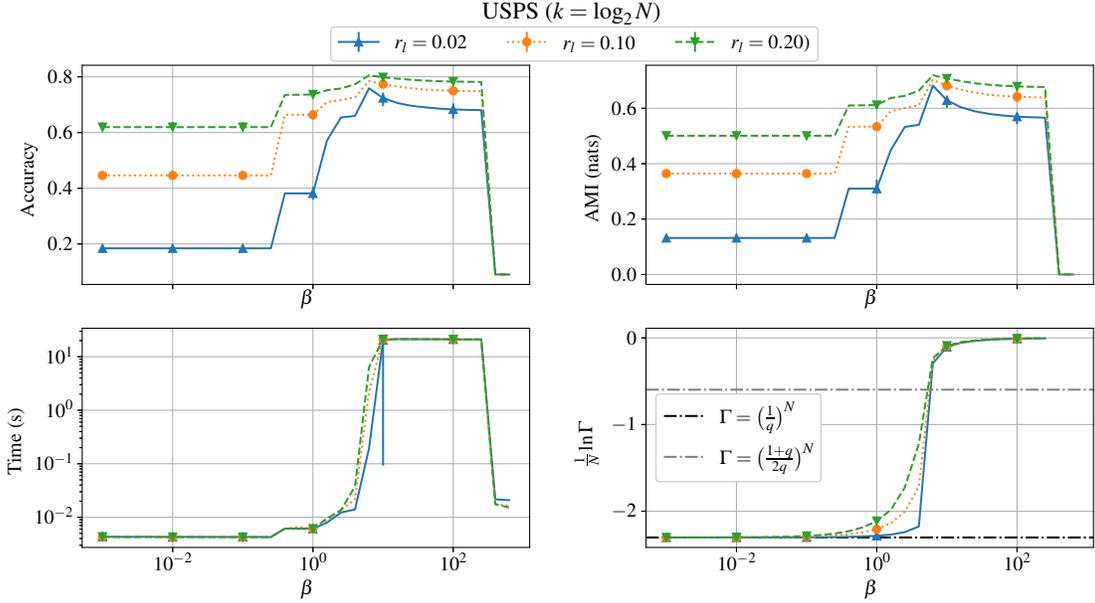}
		\caption{Results for the USPS dataset as a function of $\beta$. Lines denote averages  and bars denote maximum and minimum over different realizations of $D_{l}$. \label{fig6}}
	\end{figure}
	
	\subsection{A tuning procedure for $\beta$}
	
	The discussion made so far points out that finding the pointwise optimal value of $\beta$ is hard, specially since $\Gamma$ does not show a profile that is similar to the the accuracy and AMI curves, therefore suboptimal configurations exist at higher $\Gamma$, but those seem to be probabilistically indistinguishable from optimal ones. However, the existence of a range where classification metrics are more stable and the observation that this is associated with a plateau in $\Gamma$ indicates that one could try to tune $\beta$ by tuning $\Gamma$.  
	\par 
	The above then motivates us to find an approximation for $\Gamma$ to tune the model. We will work with the logarithm of $\Gamma$
	\begin{equation}
		\ln \Gamma = \sum_{i} (h_{i,y_{i}}-\ln Z_{i})	
	\end{equation}
	in order to make calculations easier. As noted, for low $\beta$ we have $\psi_{i}(s_{i}) \approx \frac{1}{q}$. At this regime we can also approximate $Z_{i}$ by the first terms in the Taylor series of the exponential:
	\begin{equation}
	Z_{i} \approx \sum_{s_{i}}(1+h_{i,s_{i}}) = q+\beta\bigg(\sum_{s_{i}}\theta_{i,s_{i}}+1\bigg),
	\end{equation}
	where the last expression comes from the normalization of the rows of $\mathbf{W}$ (\autoref{eq:norm}). Then,
	\begin{equation} \label{eq:gamma_approx}
		\ln \Gamma \approx \sum_{i} \Bigg[ \beta\Bigg(\theta_{i,y_{i}}+\frac{1}{q}\Bigg)-\ln\bigg\{q+\beta\bigg(\sum_{s_{i}}\theta_{i,s_{i}}+1\bigg)\bigg\}\Bigg].
	\end{equation}	 
	\par 
	Now, to evaluate the sum over all variables in the above expression we recall the definition of $\pmb{\theta}$ in \autoref{eq:fields}. Since elements in $D_{l}$ cannot change labels through NMF, summing all $\theta_{i,y_{i}}$ yields the number of elements in $D_{l}$. By the definition in \autoref{eq:fields} we also note that the sum inside the logarithm is equal to $1$ for labeled variables and $0$ for unlabeled ones. Therefore, \autoref{eq:gamma_approx} can be written as 
	\begin{equation}
		\ln \Gamma \approx \beta \Bigg( N_{l}+\frac{N}{q} \Bigg)-N_{l}\ln\{q+2\beta\}-(N-N_{l})\ln\{q+\beta\}
	\end{equation}		 
	that when divided by $N$ yields 
	\begin{equation}\label{eq:gammaN}
		\frac{1}{N} \ln\Gamma \approx \beta \Bigg( r_{l}+\frac{1}{q} \Bigg)-r_{l}\ln\{q+2\beta\}-(1-r_{l})\ln\{q+\beta\}.
	\end{equation}
	It is noteworthy that this can be used only in the low $\beta$ regime since it diverges to infinity as $\beta \rightarrow \infty$. 
	\par 
	As we have shown before based on the empirical study presented in this section, more stable results of classification are located above a threshold in $\Gamma$. Therefore we aim to choose $\beta$ by solving 
	\begin{equation}\label{eq:tuning}
		\frac{1}{N}\ln \Gamma = \ln \gamma
	\end{equation}
	for $\gamma \in (0,1]$, where $\gamma$ is an user-defined parameter. One could argue we are now simply changing the problem of choosing $\beta$ to the problem of choosing $\gamma$, which is true. Choosing $\gamma$, however, points to choosing how sure one wants to be about a classification, with the true level of belief being given by the actual value of $\Gamma$. This does not indicate that a higher $\gamma$ will leads to better classification results, but, together with the aforementioned stability at higher $\Gamma$, this procedure has a foundation for its applicability.
	\par
\begin{figure}[h]
		\centering
		\includegraphics[scale=0.6]{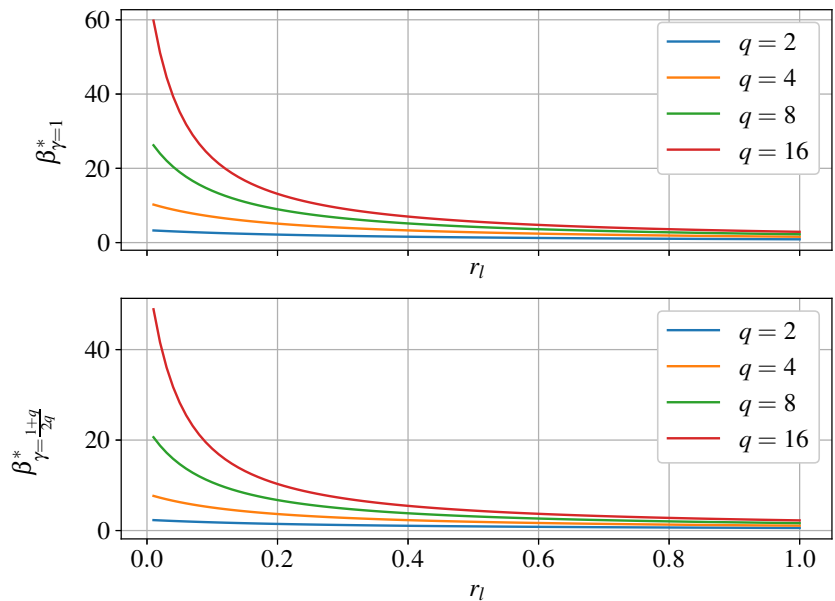}
		\caption{Solution of \autoref{eq:tuning} as a function of $r_{l}$ for two different values of $\gamma$ and four different values of $q$. \label{fig7}} 
	\end{figure}

 
	\autoref{fig7} shows the solution $\beta^{*}_{\gamma}$ of \autoref{eq:tuning} using the approximation in \autoref{eq:gammaN} and setting $\gamma = \frac{1+q}{2q}$ and $\gamma = 1$ as target values. To solve the non-linear equation we used the implementation of Newton's method in the SciPy library \cite{virtanen2020} with initial guess set to unity and tolerance set to $10^{-3}$.
	\par 
	As \autoref{eq:gammaN} is designed to work at lower values of $\beta$, one may face a problem in its correctness in situations where the number of classes $q$ is too high or when the amount of labeled data provided is too low, since these conditions increase the values of $\beta^{*}_{\gamma}$ (\autoref{fig7}). 
	\par 
	Now, the existence of a procedure for tuning $\beta$ will allow us to make a more in-depth comparison of NMF with GRF and LGC. In the next section we will use both values of $\gamma$ to calculate $\beta^{*}_{\gamma}$ and compare these approaches with the classical approaches in the field of semi-supervised learning. 
	
\section{Comparative evaluation of SSL algorithms}

	In this section we evaluate GRF, LGC and NMF using the previously constructed tuning approach for $\beta$ with $\gamma = \frac{1+q}{2q}$ and $\gamma=1$. The first is the observed lower bound reported in the previous section, while the second should be the limit for $\gamma$ where our approximation holds. For all algorithms we set $t_{max}=10^{4}$ and $\epsilon = 10^{-3}$, while for LGC we set $\alpha = 0.99$ as in the original paper \cite{zhou2004}. Experiments are then conducted for fixed values of $k$ and $r_{l}$ while varying the other quantity.
	
	\subsection{Bidimensional datasets}
	
		\begin{figure}[h]
		\centering
		\includegraphics[scale=0.5]{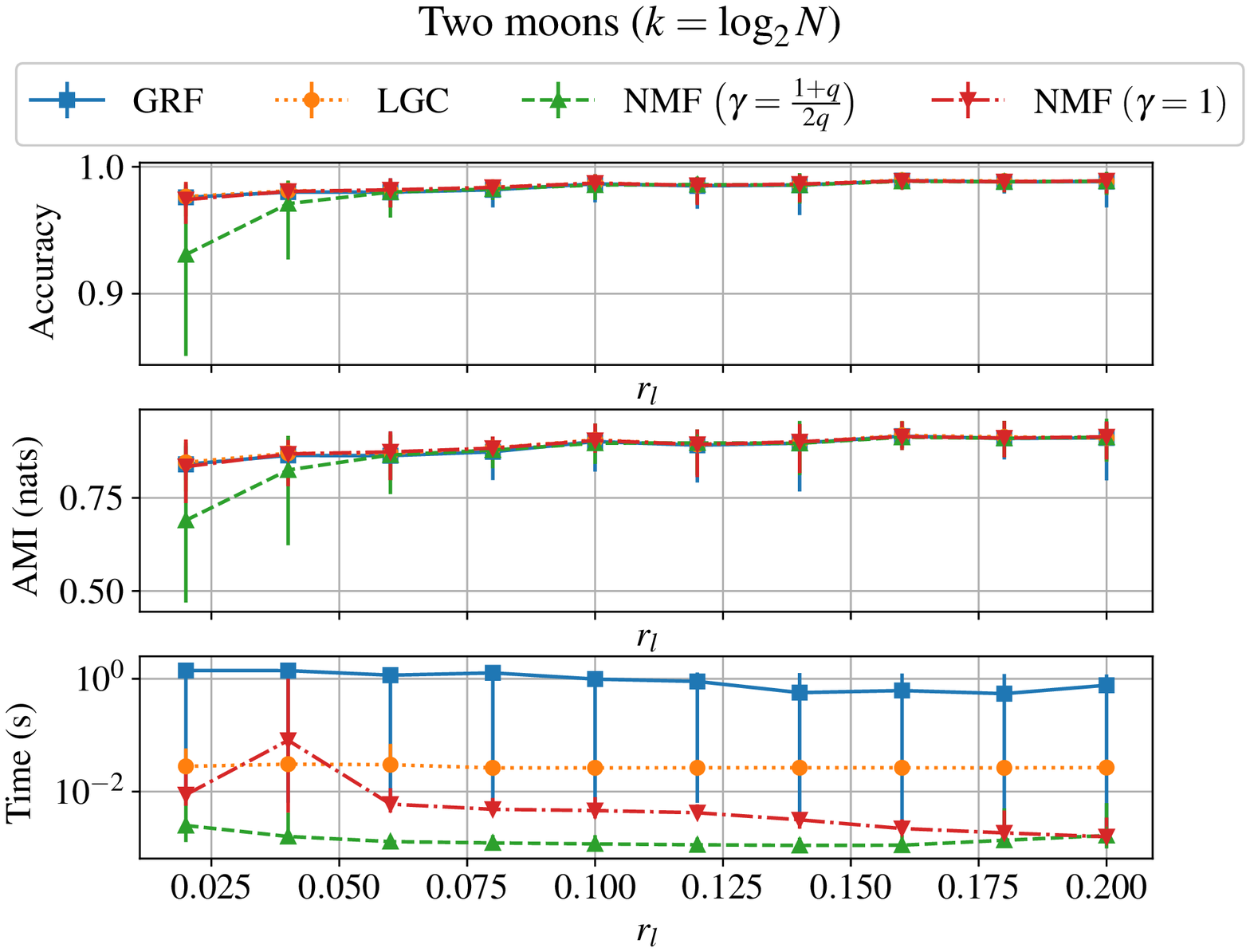}
		\includegraphics[scale=0.5]{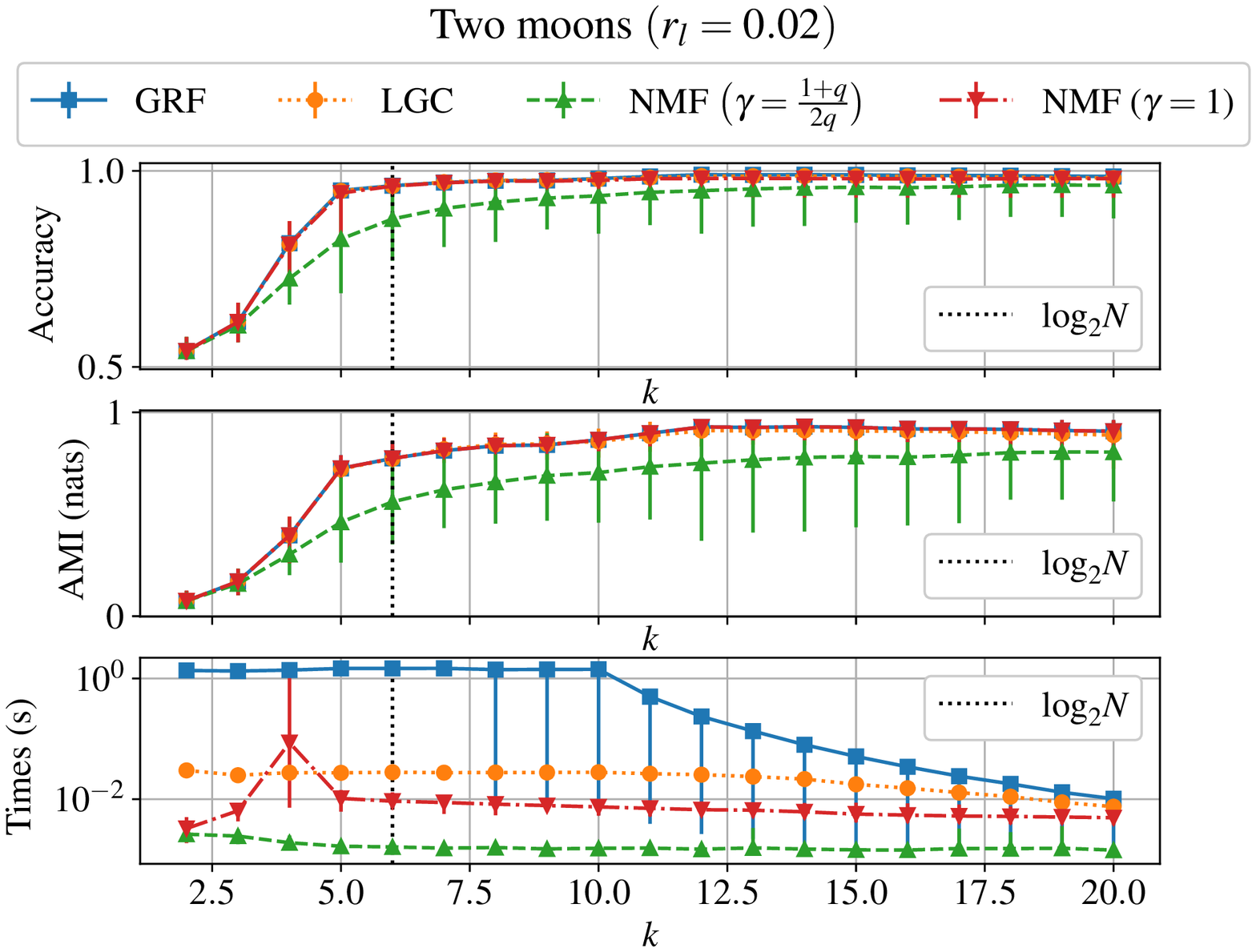}
		\caption{Results for GRF, LGC and NMF as functions of $k$ and $r_{l}$. Lines denote averages and bars denote maximum and minimum over different realizations of $D_{l}$. \label{fig8}}
	\end{figure}

	\begin{figure}[h]
		\centering
		\includegraphics[scale=0.5]{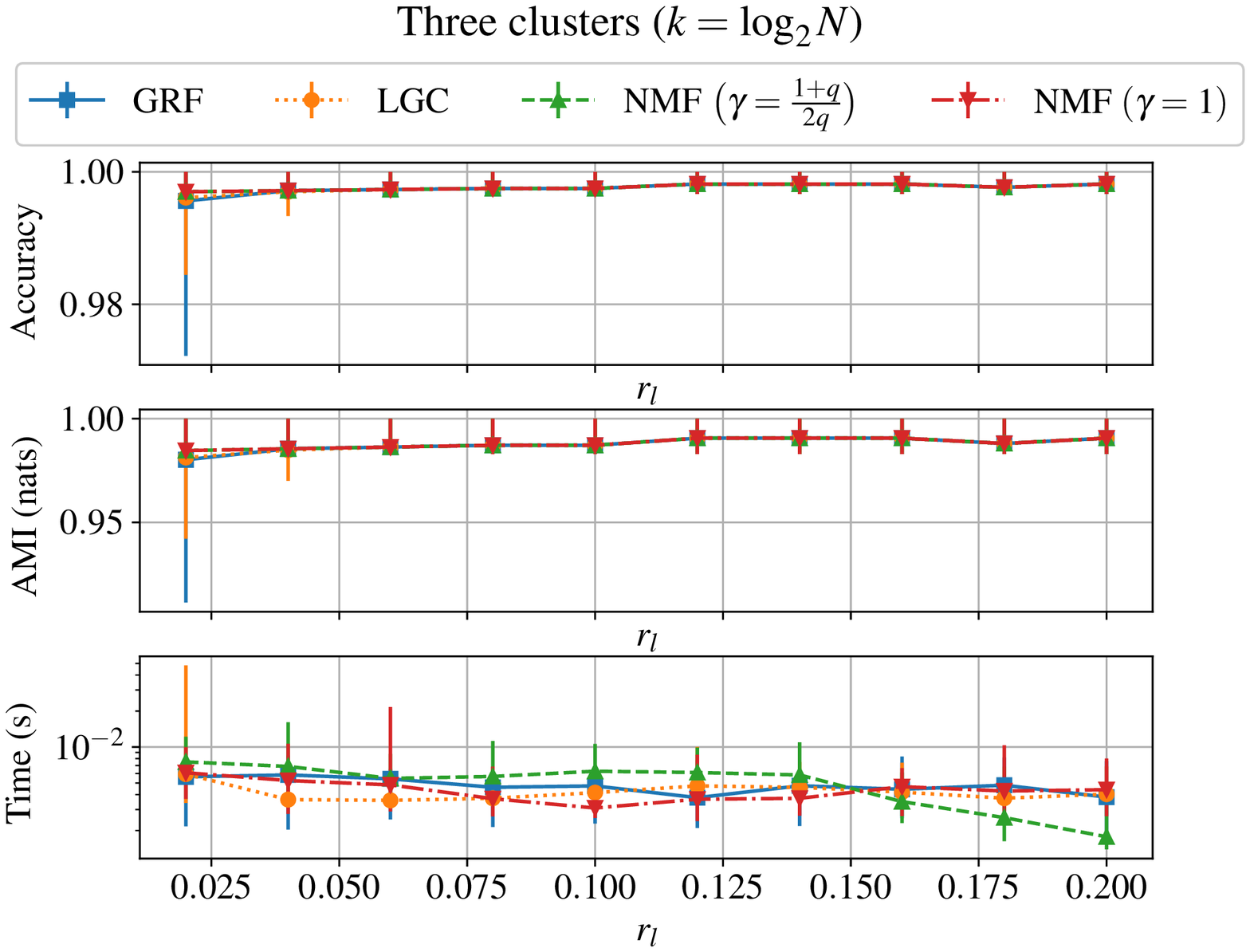}
		\includegraphics[scale=0.5]{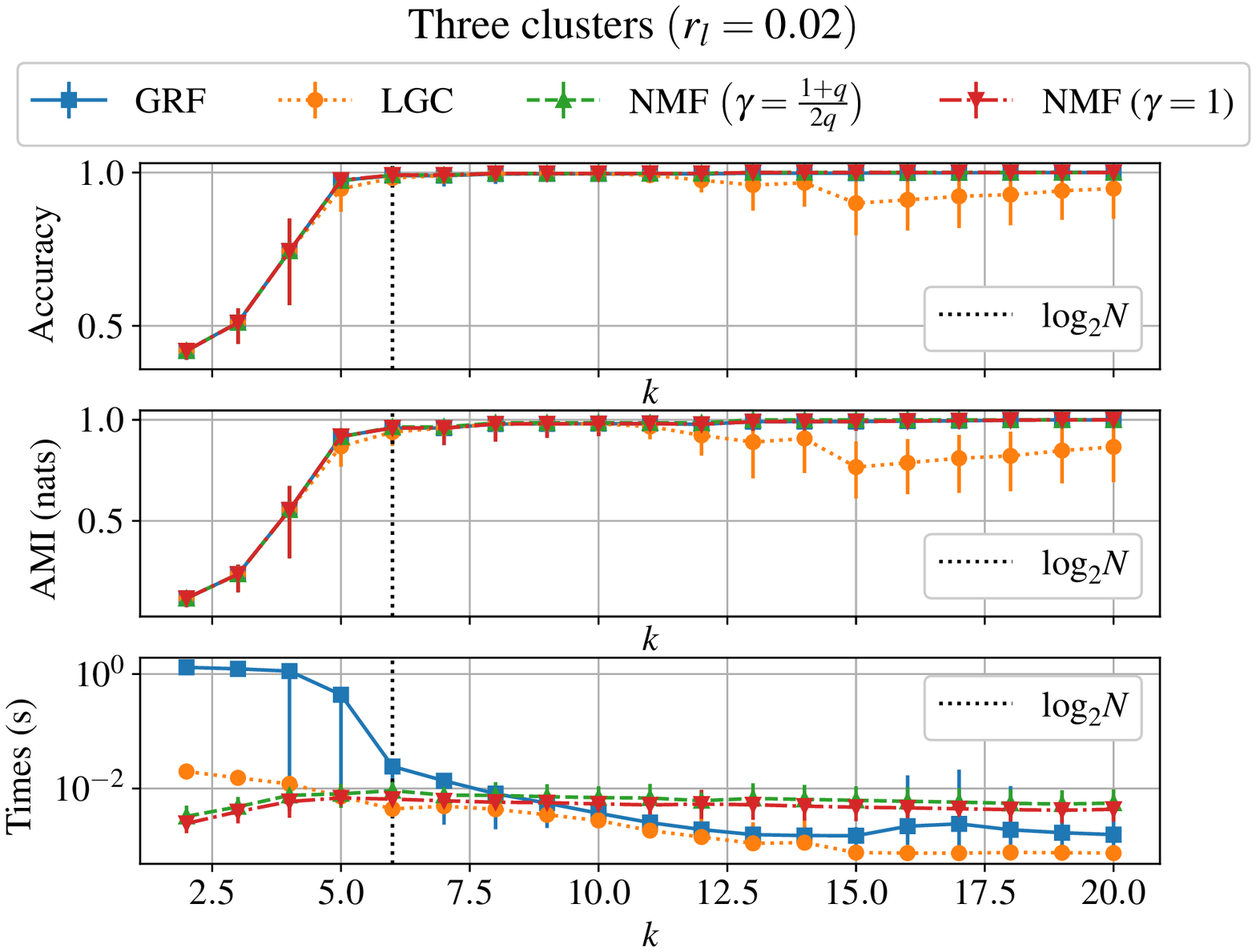}
		\caption{Results for GRF, LGC and NMF as functions of $k$ and $r_{l}$. Lines denote averages and bars denote maximum and minimum over different realizations of $D_{l}$. \label{fig9}}
	\end{figure}
	
	\begin{figure}[h]
		\centering
		\includegraphics[scale=0.5]{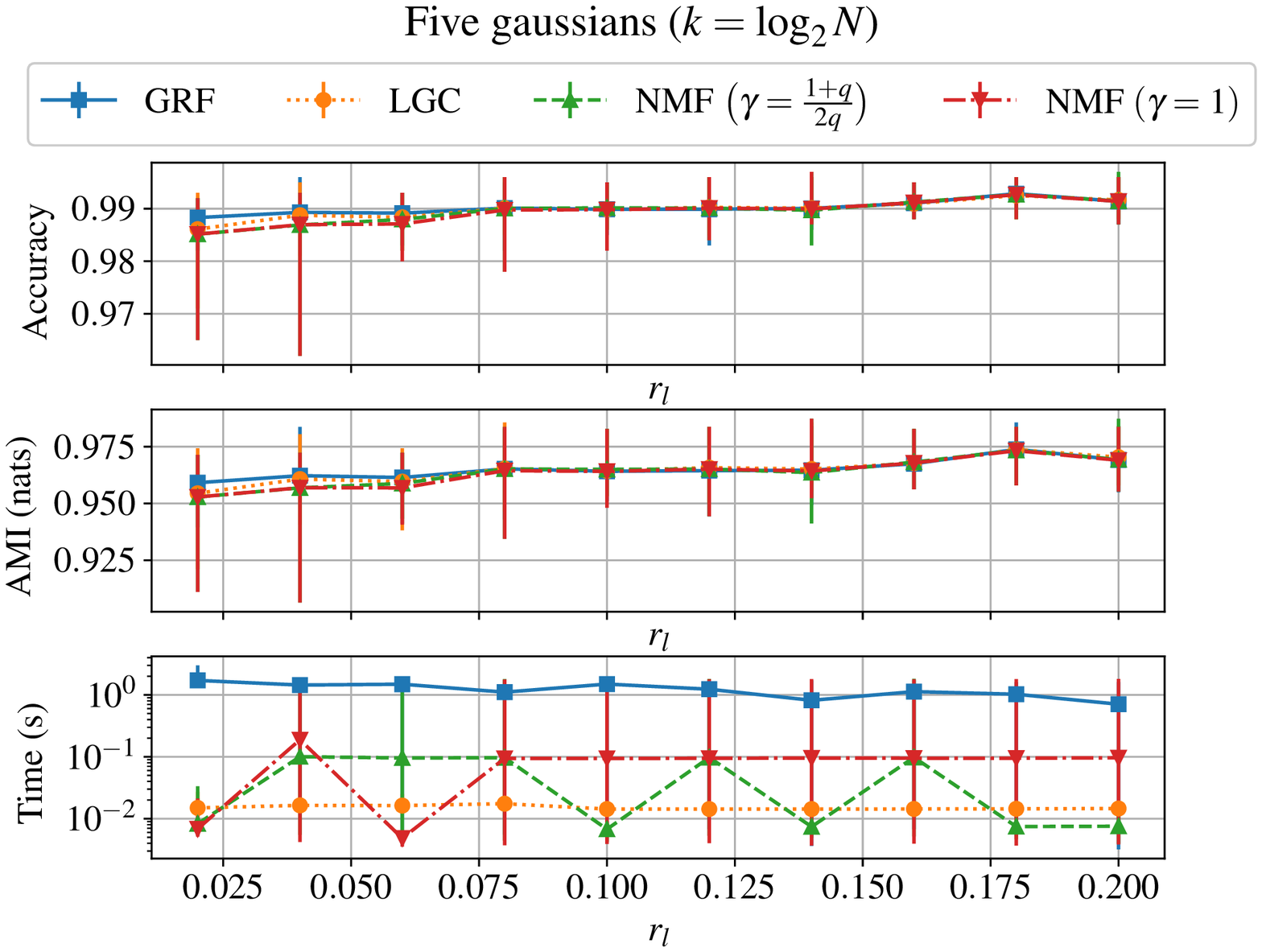}
		\includegraphics[scale=0.5]{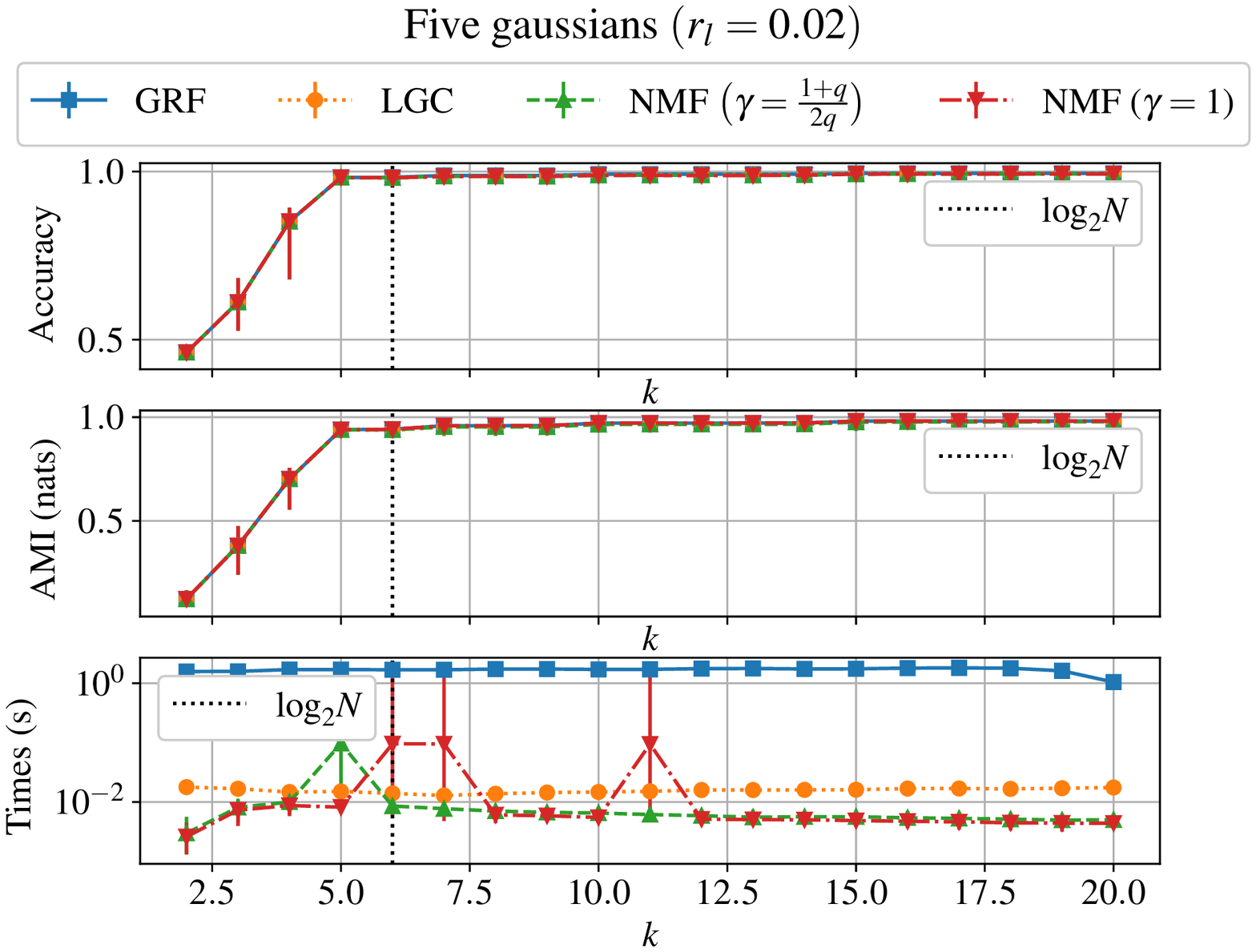}
		\caption{Results for GRF, LGC and NMF as functions of $k$ and $r_{l}$. Lines denote averages and bars denote maximum and minimum over different realizations of $D_{l}$. \label{fig10}}
	\end{figure}
	
	As shown in figures 8-10 the behavior of the three algorithms is highly determined by the datasets, which makes it difficult to point out the most effective approach.
	\par 
	In the two moons dataset (\autoref{fig8}) setting $\gamma=\frac{1+q}{2q}$ provided a faster execution time at the expense of accuracy and AMI, especially in the regime of small $r_{l}$. The approach for $\gamma=1$ also showed a better computational performance than GRF and LGC in most studied scenarios, but without the cost in classification metrics shown by the other tunning approach.
	\par 
	When looking at accuracy and AMI as functions of $k$ in \autoref{fig8} one sees that the difference in the two ways of choosing $\beta$ is related to the first being less susceptible to increases in $k$. The second method has similar behavior to GRF and LGC in terms of classification, while being faster than both of these methods, especially for lower $k$.
	\par 
	The case of three clusters (\autoref{fig9}) showed to favor NMF approaches by a small portion in the low $r_{l}$ regime. However, in denser topologies, GRF and LGC become faster, but the second shows a decrease in classification metrics for $k>11$. 
	\par 
	NMF shows an irregular execution time profile as a function of $r_{l}$ in the five gaussians dataset, as illustrated in \autoref{fig10}. Since the procedure for choosing $\beta$ depends strictly on $r_{l}$ but not on the particular realizations of $D_{l}$, we see that the classification metrics of NMF can be highly affected by different configurations of the previously labeled instances of the dataset, particularly as labeled data becomes more scarce.
	\par 
	What we note as the similarity between the three studied datasets is that the increase in $k$ tends to, on average, improve the execution time of the algorithms for $k>\log_{2}N$, with the exception of LGC in the five gaussians dataset (\autoref{fig10}). This is quite interesting behavior as one could expect that the addition of edges to $G_{D}$ would slow down these algorithms. We believe this is related to the procedure of symmetrization and sparsification discussed in \autoref{subsec22}, which allows for the increase in $k$ to connect more similar points and increase intra-class connectivity, making the problem easier to handle for the algorithms and enhancing their convergence.
	\par 
	The above is supported by the average improvement in classification metrics as $k$ increases. The exception is again LGC, but in the three clusters dataset (\autoref{fig9}). Therefore our study so far leads us to believe that LGC (with $\alpha=0.99$) is affected by changes in the topology of $G_{D}$ in a different manner than GRF and NMF, which is an expected behavior based on a previous study that connects the Potts model with GRF \cite{tibely2008}.
	\par 
	We also highlight that in the bidimensional datasets (Figures 8-10) the accuracy and AMI curves show a high agreement: they respond similarly to variations in $r_{l}$ and $k$. Therefore, classification and clustering using GRF, LGC and NMF are closely related in the said datasets.

	\subsection{High-dimensional datasets}
	
	\begin{figure}[h]
		\centering
		\includegraphics[scale=0.5]{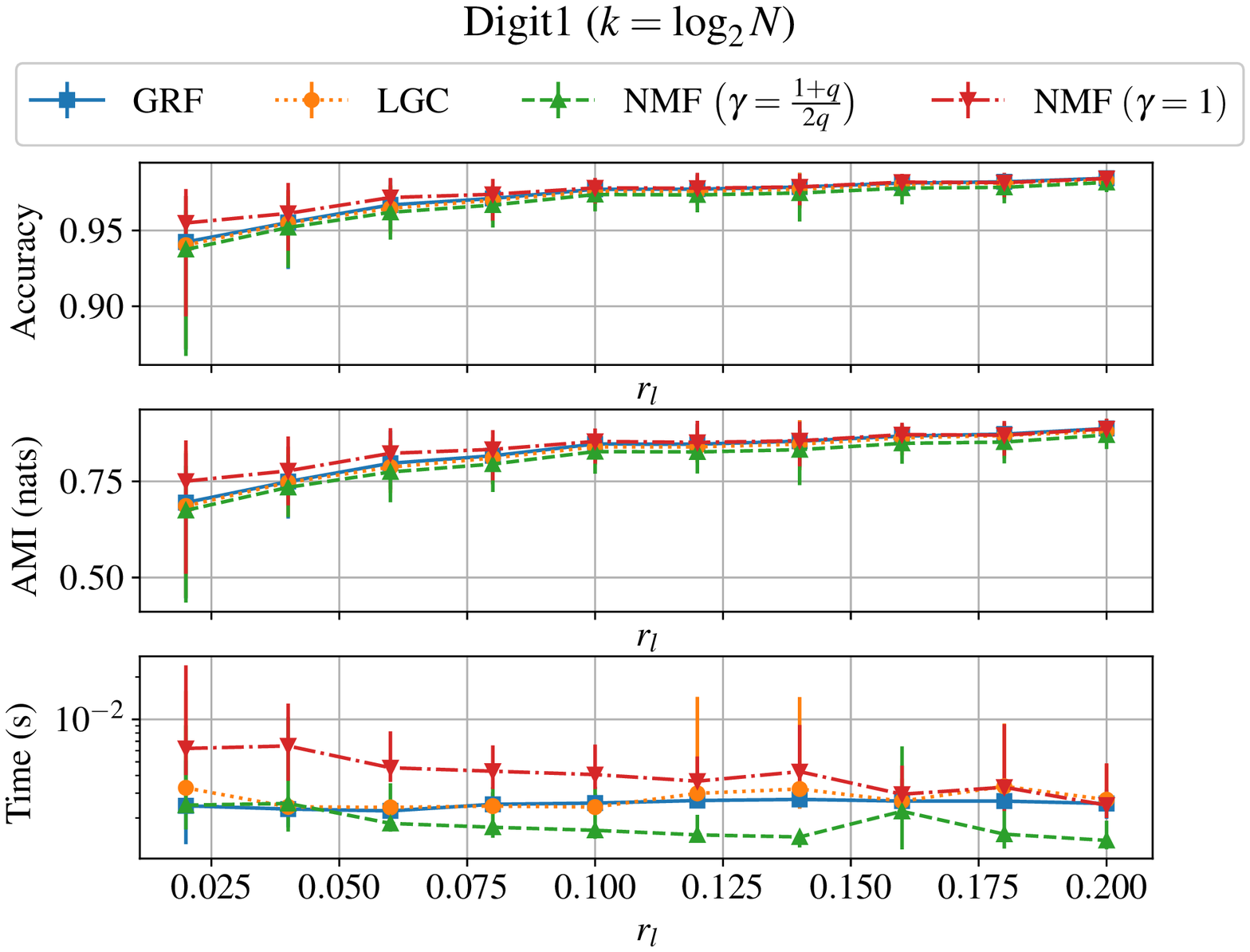}
		\includegraphics[scale=0.5]{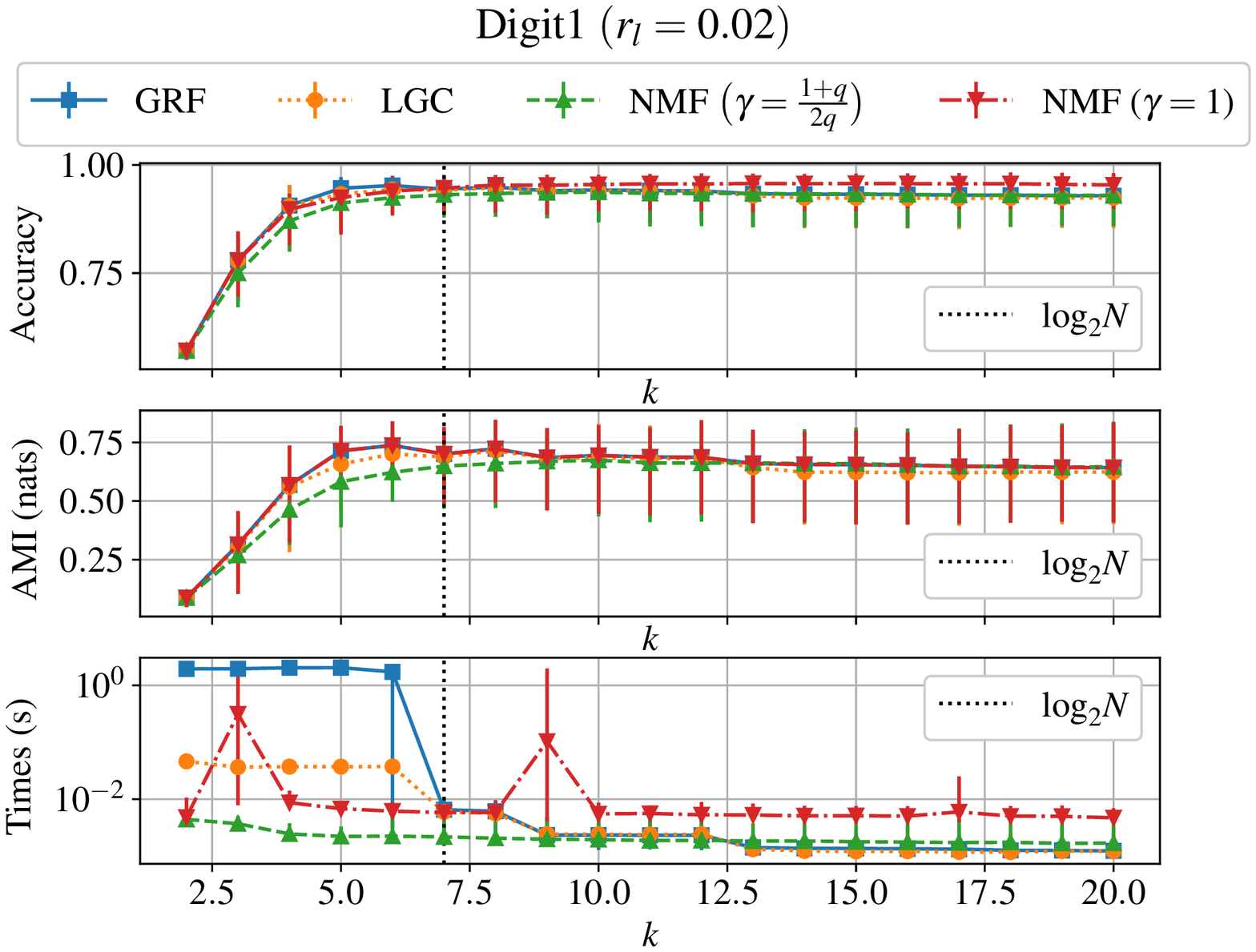}
		\caption{Results for GRF, LGC and NMF as functions of $k$ and $r_{l}$. Lines denote averages and bars denote maximum and minimum over different realizations of $D_{l}$. \label{fig11}}
	\end{figure}
	
	\begin{figure}[h]
		\centering
		\includegraphics[scale=0.5]{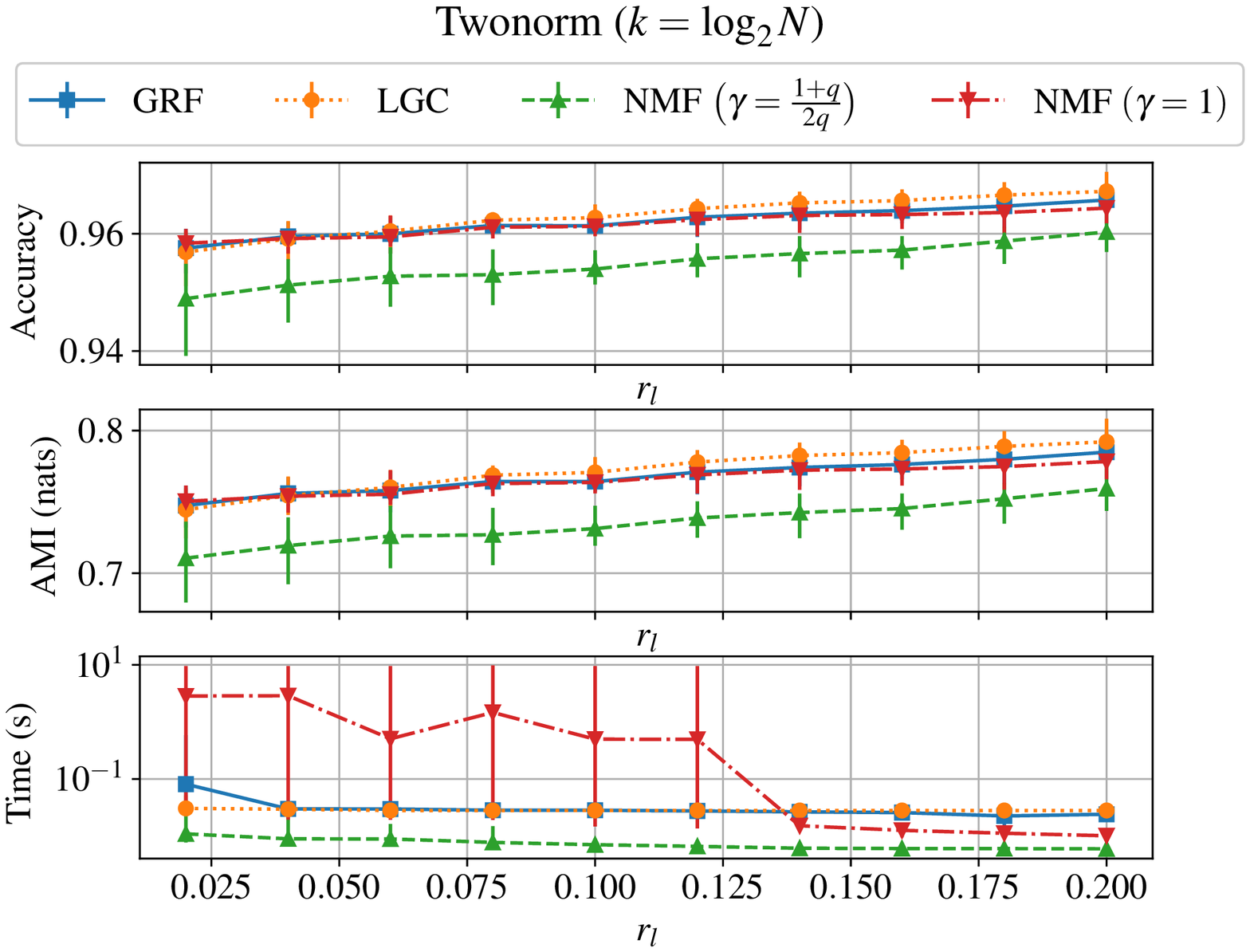}
		\includegraphics[scale=0.5]{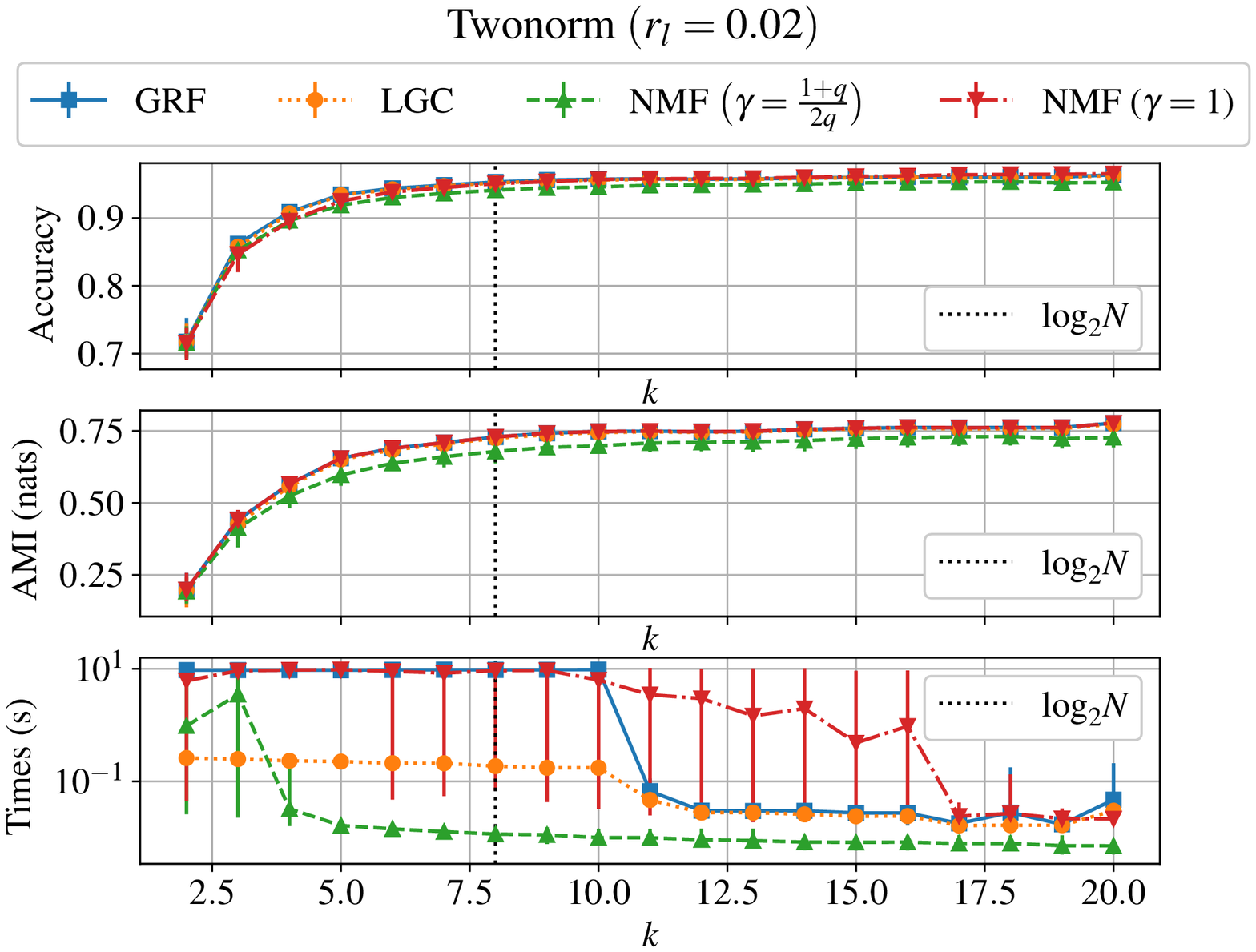}
		\caption{Results for GRF, LGC and NMF as functions of $k$ and $r_{l}$. Lines denote averages and bars denote maximum and minimum over different realizations of $D_{l}$. \label{fig12}}
	\end{figure}
	
	\begin{figure}[h]
		\centering
		\includegraphics[scale=0.5]{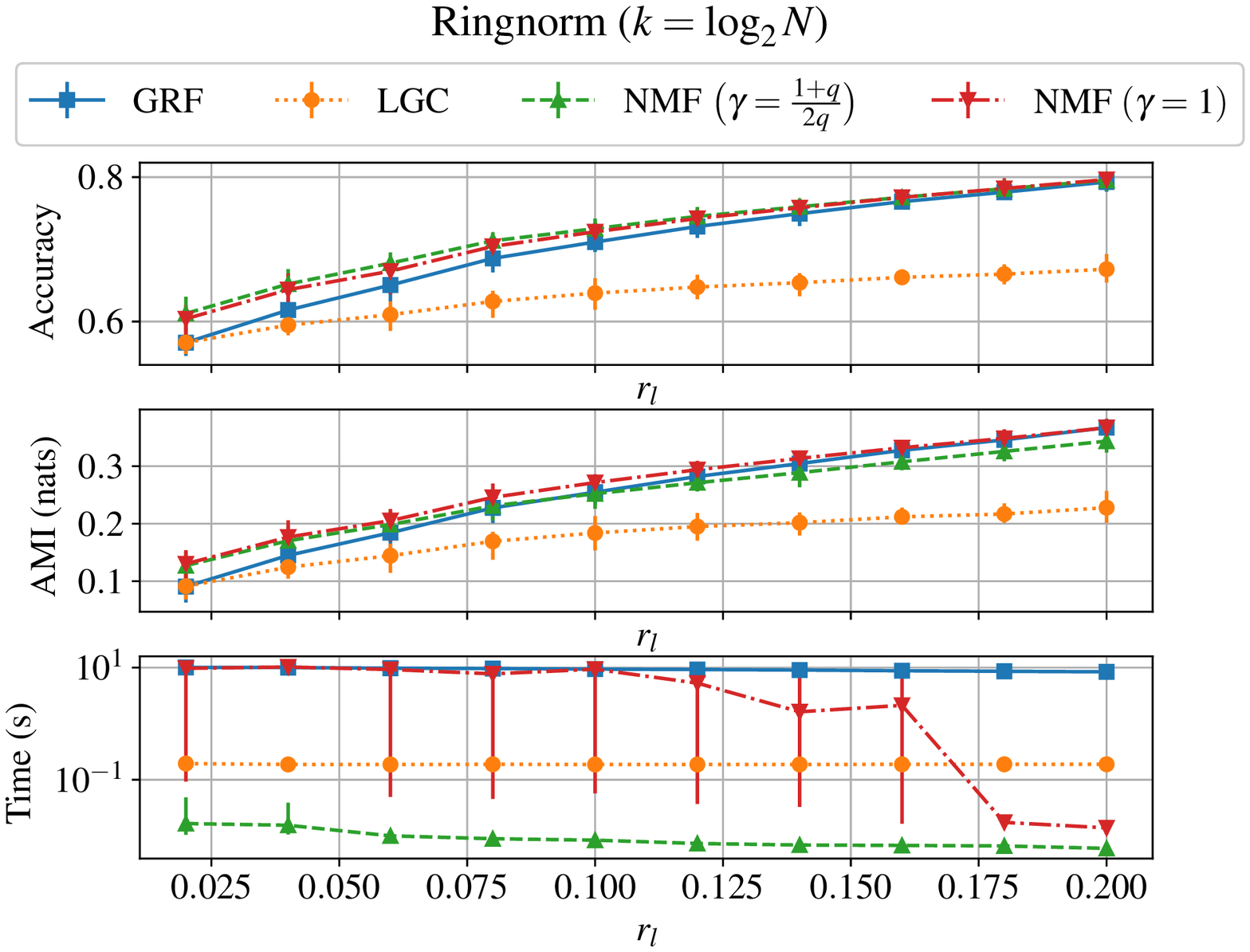}
		\includegraphics[scale=0.5]{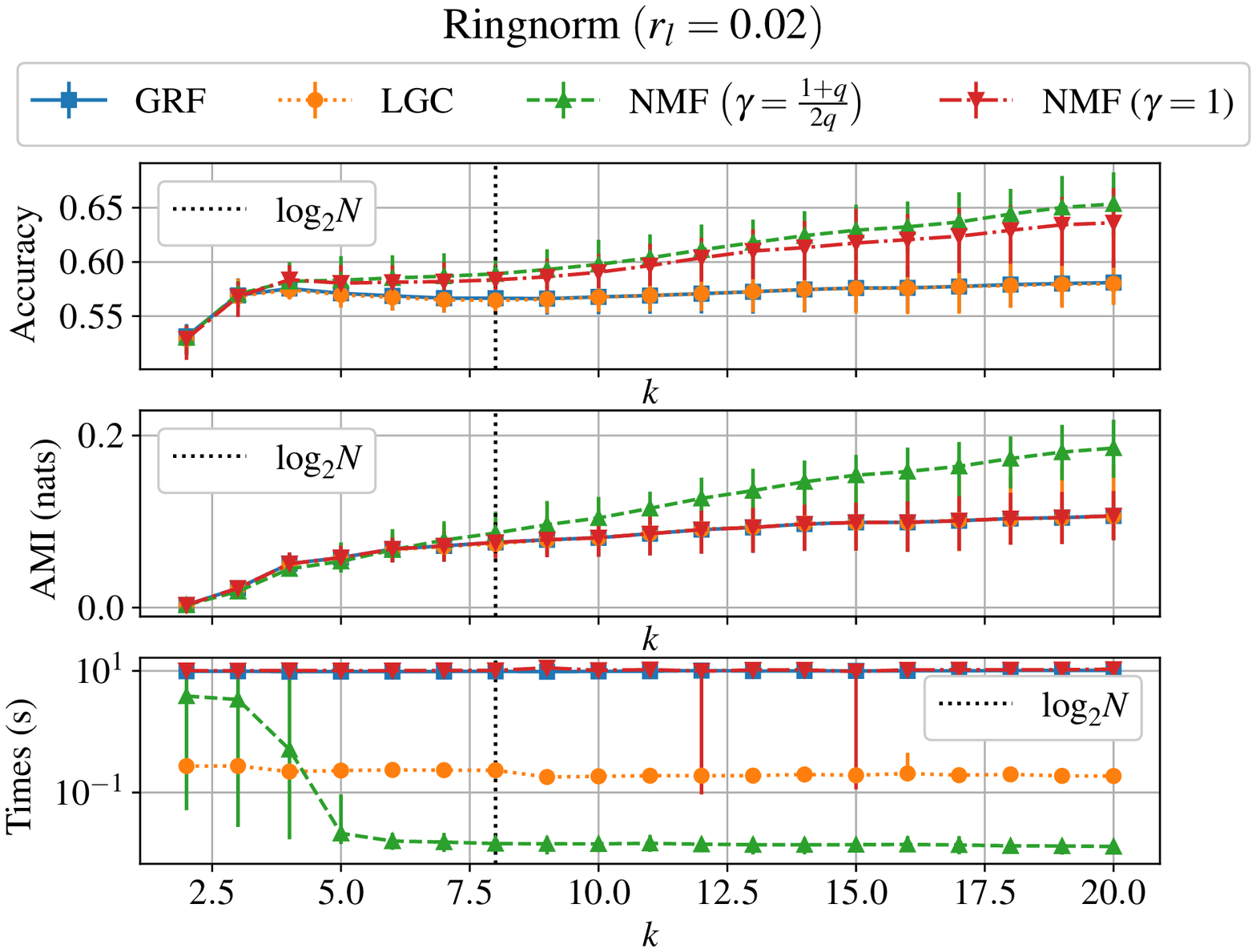}
		\caption{Results for GRF, LGC and NMF as functions of $k$ and $r_{l}$. Lines denote averages and bars denote maximum and minimum over different realizations of $D_{l}$. \label{fig13}}
	\end{figure}
	
	\begin{figure}[h]
		\centering
		\includegraphics[scale=0.5]{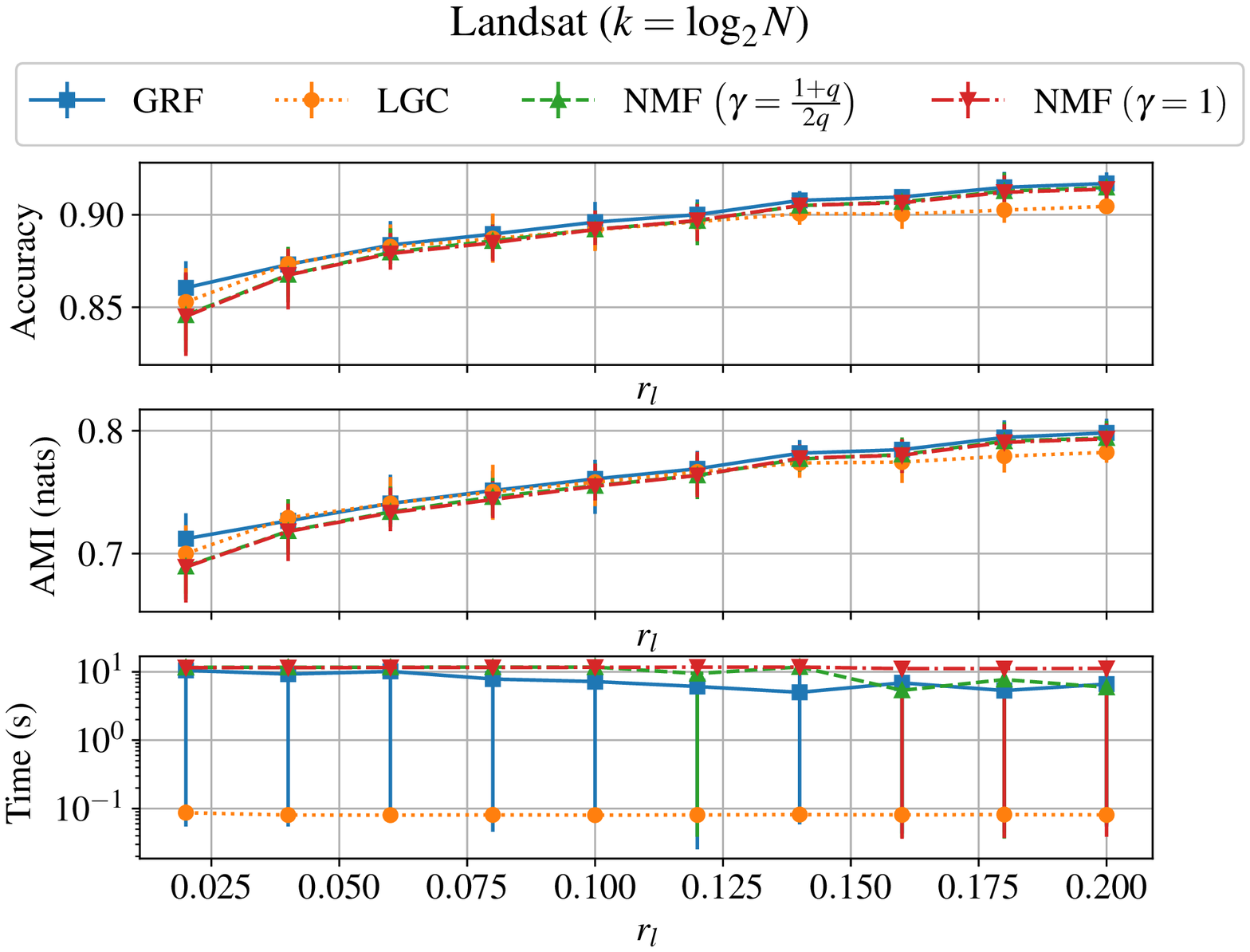}
		\includegraphics[scale=0.5]{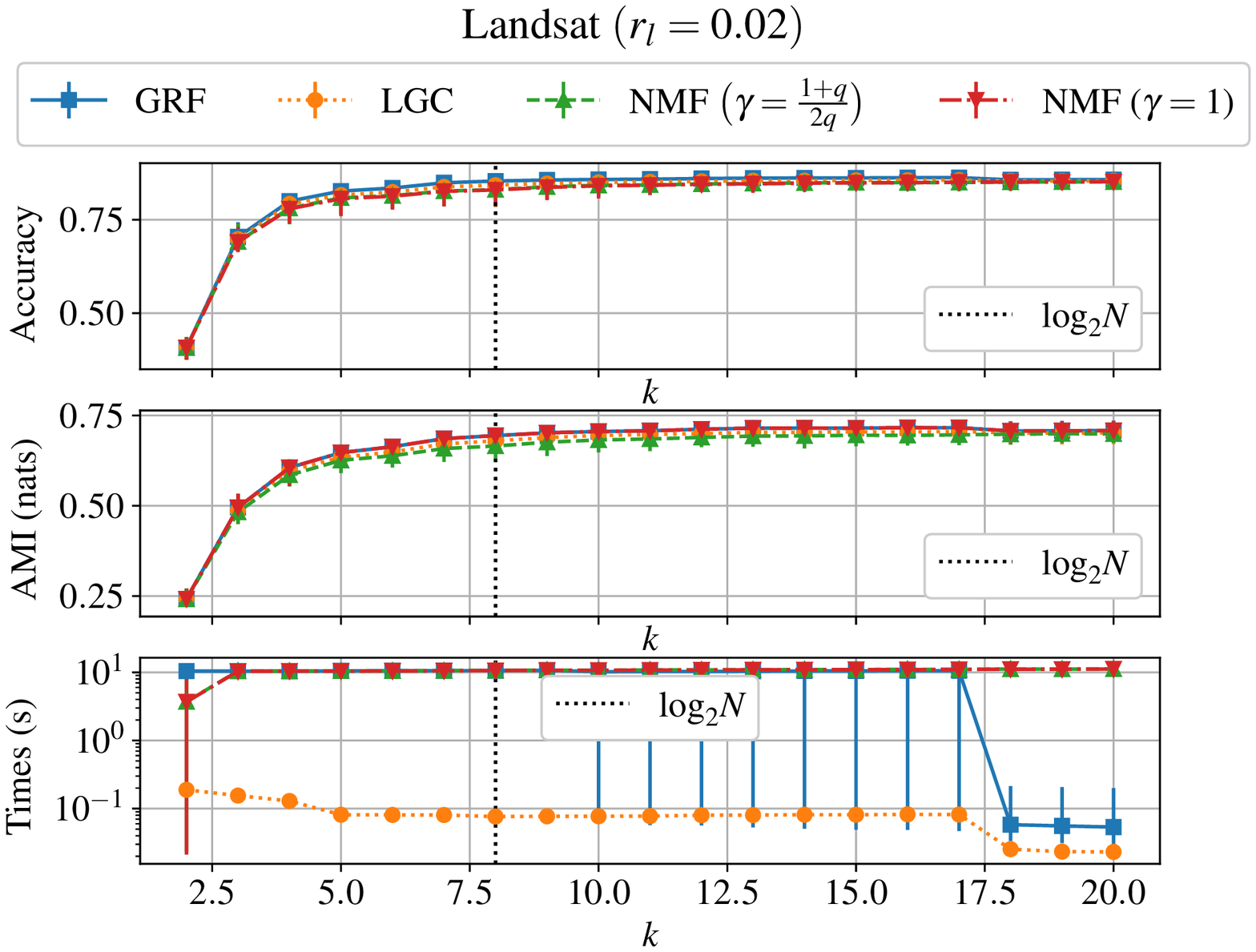}
		\caption{Results for GRF, LGC and NMF as functions of $k$ and $r_{l}$. Lines denote averages and bars denote maximum and minimum over different realizations of $D_{l}$. \label{fig14}}
	\end{figure}
	
	\begin{figure}[h]
		\centering
		\includegraphics[scale=0.5]{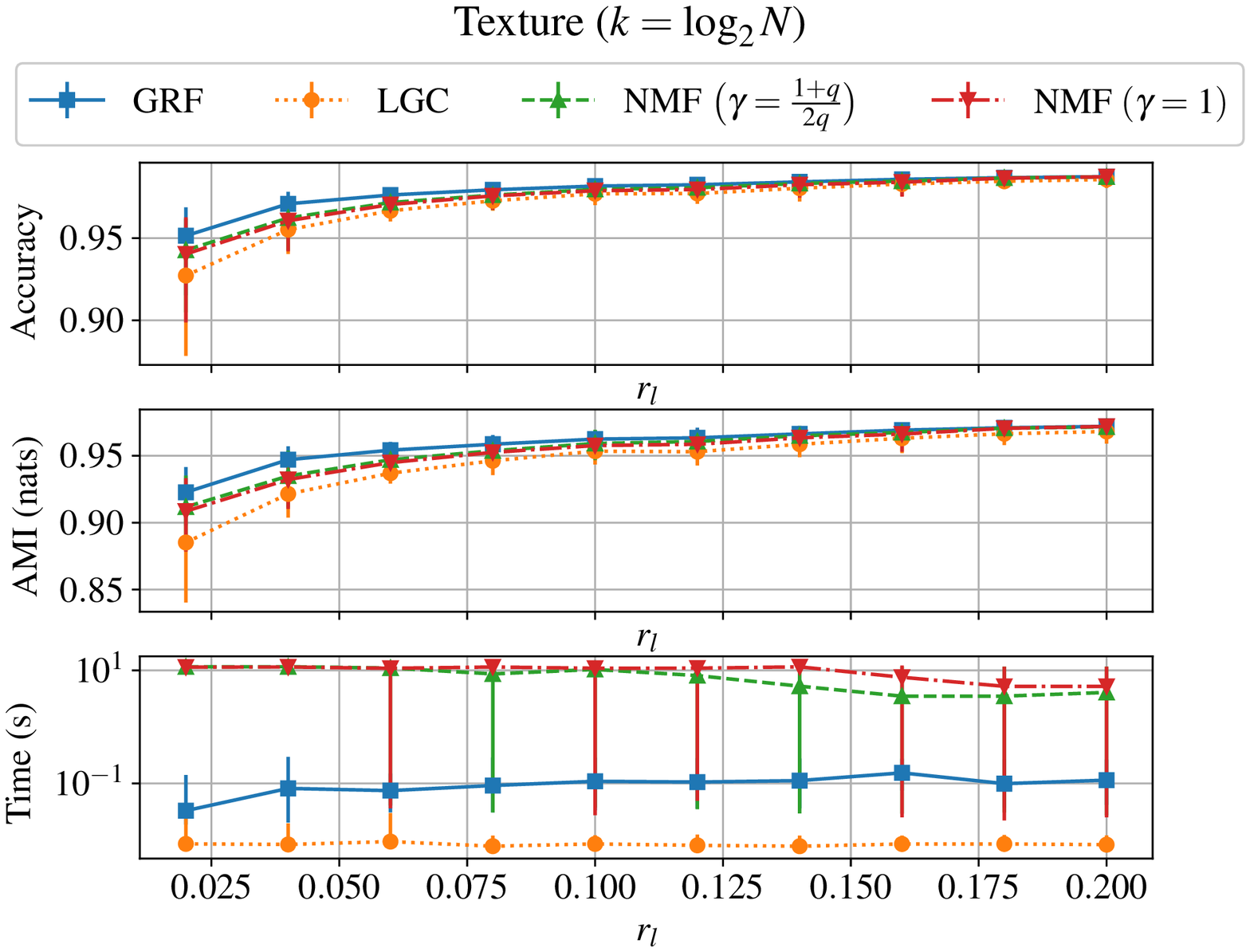}
		\includegraphics[scale=0.5]{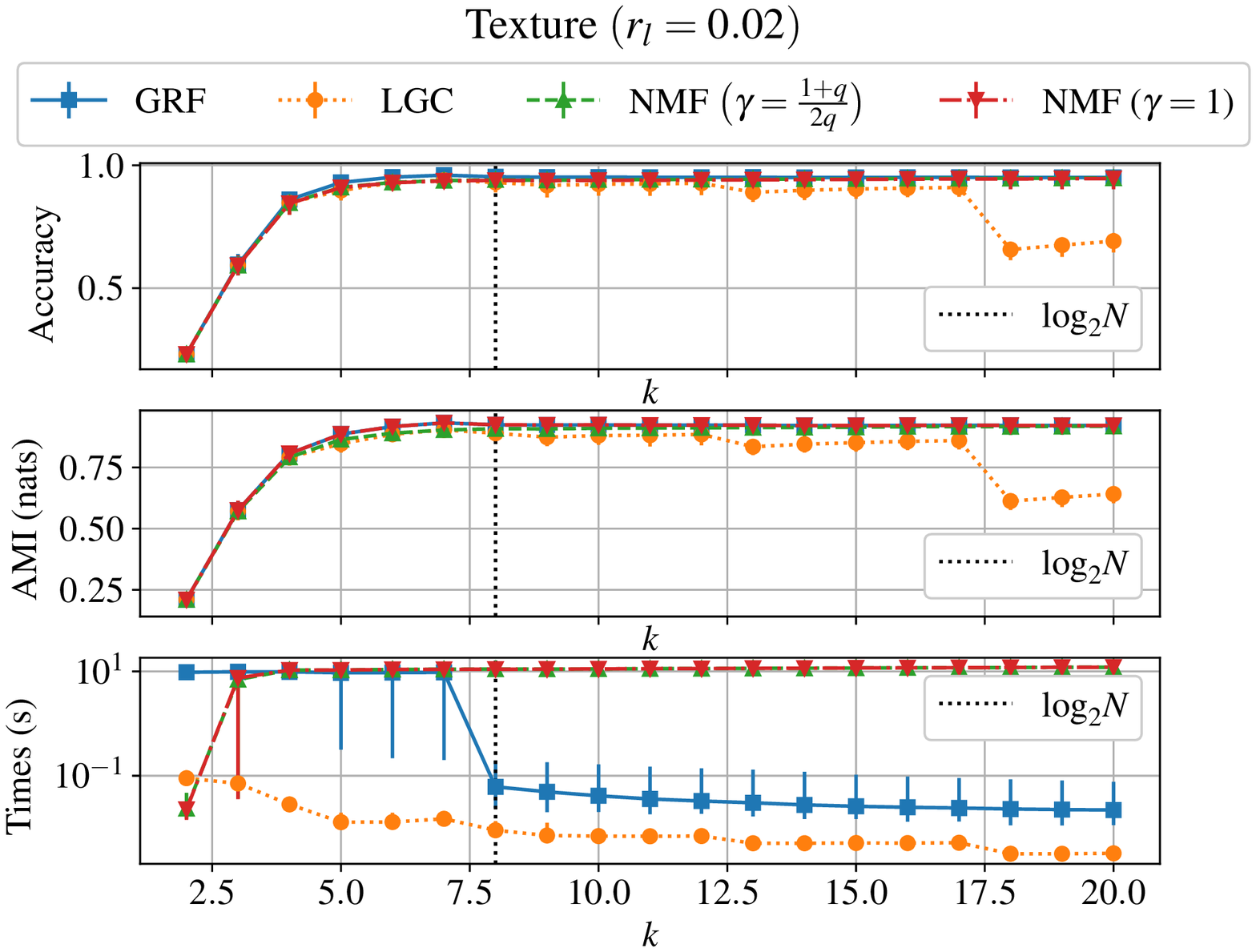}
		\caption{Results for GRF, LGC and NMF as functions of $k$ and $r_{l}$. Lines denote averages and bars denote maximum and minimum over different realizations of $D_{l}$. \label{fig15}}
	\end{figure}
	
	\begin{figure}[h]
		\centering
		\includegraphics[scale=0.5]{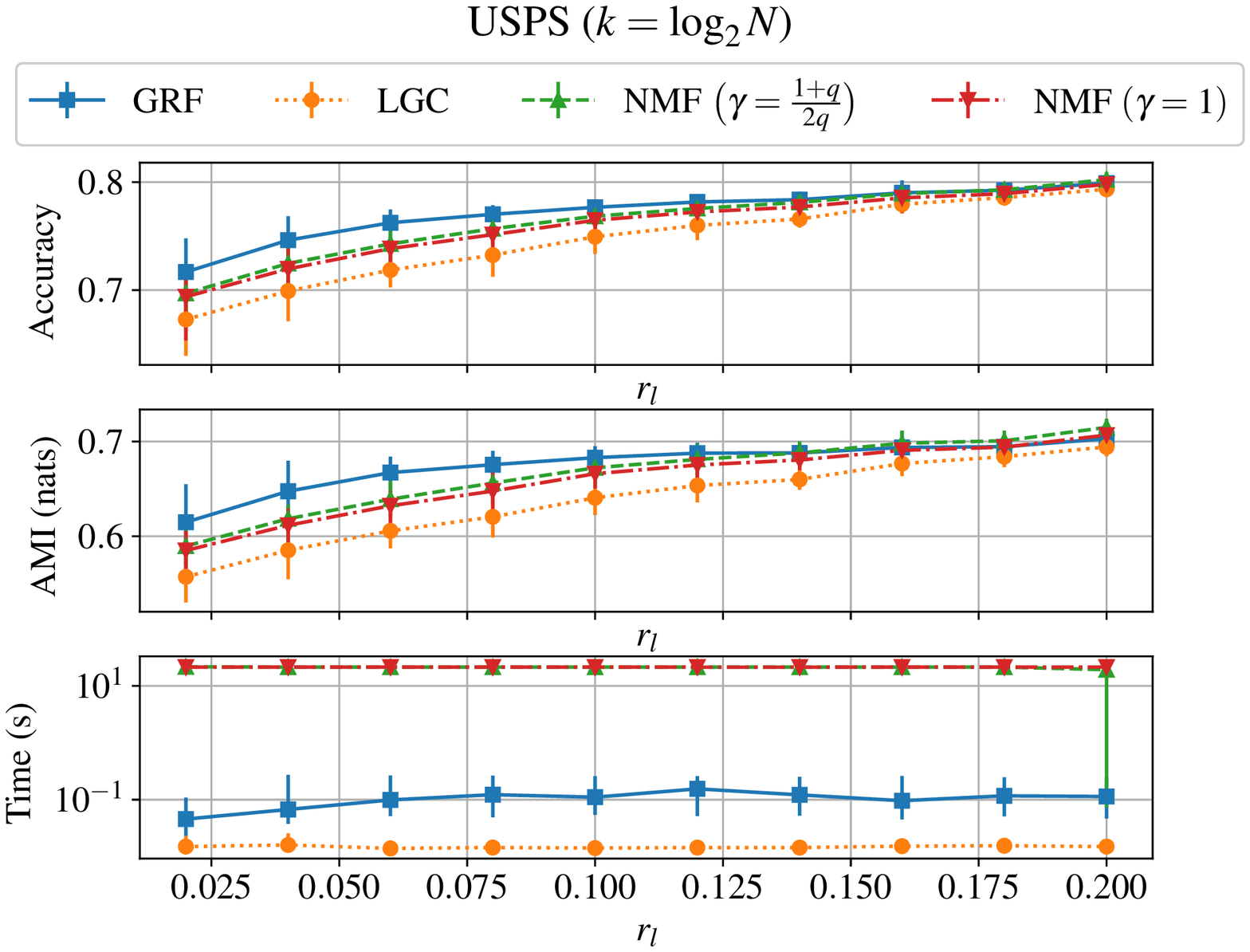}
		\includegraphics[scale=0.5]{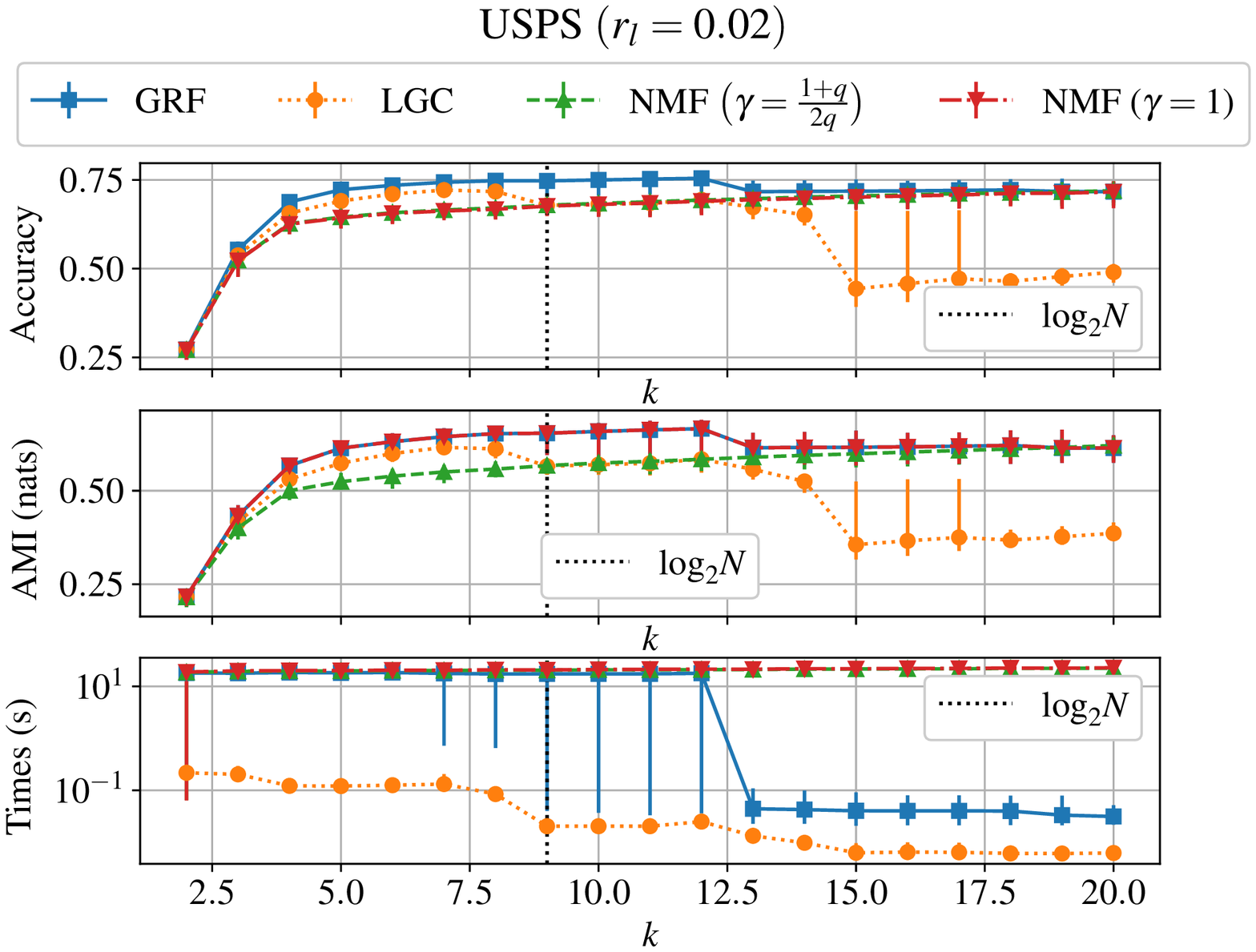}
		\caption{Results for GRF, LGC and NMF as functions of $k$ and $r_{l}$. Lines denote averages and bars denote maximum and minimum over different realizations of $D_{l}$. \label{fig16}}
	\end{figure}
	
	As we now move to analyze high-dimensional datasets we will also investigate data that comes from real-world phenomena, such as Landsat, Texture and USPS. Our discussion will navigate by each of the six sets being analyzed.
	\par 
	For Digit1 (\autoref{fig11}) we observe the utilization of NMF with $\gamma=1$ yields the best results in accuracy and AMI when the problem is looked at as a function of $r_{l}$. When we analyze the $k$ dependency of the results we start observing some differences compared to the bidimensional case: a divergence between accuracy and AMI appears (particularly for $k=5$ and $k=6$), since $\gamma=1$ outperforms GRF in the first metric but not in the second. As $k$ increases past $\log_{2}N$, AMI decreases for all algorithms and is followed in a less pronounced way by the accuracy.
	\par 	
	Regarding computational performance of Digit1 (\autoref{fig11}), for $k<\log_{2}N$ there is some irregularity in execution time, with the exception of NMF with $\gamma = \frac{1+q}{2q}$. With $k\geq \log_{2}N$ algorithms stabilize and have similar performances.
	\par 
	Results on the Twonorm (\autoref{fig12}) dataset are closer to what is expected, based on the discussion on bidimensional datasets. NMF with $\gamma=1$, LGC and GRF have similar behavior in their accuracy and AMI curves as a function of $r_{l}$. The accuracy of $\gamma=\frac{1+q}{2q}$ is only slightly worse than the other algorithms in absolute terms, while AMI for this method is around $0.04$ smaller than other methods. However, this approach showed to be consistently faster both as a function of $r_{l}$ and as a function of $k$. In fact, when accuracy and AMI are analyzed as functions of $k$, the difference between methods becomes even less relevant. 
	\par 
	On Ringnorm (\autoref{fig13}) setting $\gamma=\frac{1+q}{2q}$ produces the most accurate results. In situations where labeled data is more scarce, this method also scores the higher values for AMI, but as $r_{l}$ increases it is overcome by GRF and $\gamma=1$. However, when we consider different constructions of $G_{D}$, both NMF approaches are more accurate than GRF and LGC by a significant margin when $k\geq \log_{2}N$. In the same range, $\gamma=\frac{1+q}{2q}$ has a higher AMI than every other method and is also faster by at least one order of magnitude.
	\par 
	Landsat (\autoref{fig14}) is a more well-behaved case regarding accuracy and AMI, with all methods showing very close results. We note that LGC seems to become less susceptible to the addition of labeled data as $r_{l}>0.15$. It is also noteworthy that LGC is the fastest approach in this dataset by two orders of magnitude, with exception of $k>17$ where GRF has a significant improvement in execution time. 
	\par 
	Experiments in the Texture dataset (\autoref{fig15}) showed some divergence among methods in the region of lower $r_{l}$, with GRF being the better method regarding accuracy and AMI, followed by NMF and then LGC. However, increasing the amount of labeled data makes the algorithms indistinguishable regarding those metrics. We also note that LGC and GRF are around two orders of magnitude faster than NMF for $k\geq \log_{2}N$. 
	\par 
	When we look at the results as a function of $k$ for Texture, LGC shows the same behavior we highlighted for the three clusters dataset (\autoref{fig9}). For $k>\log_{2}N$ accuracy and AMI suffer a significant drop, indicating that a denser graph may blur its ability to distinguish between different classes and clusters. 
	\par 
	Finally, for USPS (\autoref{fig16}) NMF is again the slower method, followed once more by GRF and LGC. Accuracy and AMI results as a function of $r_{l}$ behave similarly to the Texture dataset (\autoref{fig15}), but the divergence between methods fades at higher values of $r_{l}$. 
	\par 
	When we analyze results in USPS as a function of $k$ we observe a very rich behavior by all algorithms. Both NMF approaches show basically the same accuracy curve, while the case $\gamma=1$ is much more similar to GRF in terms of AMI. In fact, these two approaches show a dip in AMI for $k\geq\log_{2}N$, while the accuracy of NMF, in this case, is smoother. LGC, on the other hand, shows two different falls in accuracy and LGC at different values of $k$, a phenomenon that also occurs in the execution time, as in GRF. The case of NMF with $\gamma=\frac{1+q}{2q}$ draws attention in this dataset due to its monotonicity with respect to $k$, while other methods are more susceptible to different constructions of $G_{D}$. 
	
	\subsection{Discussion}
	
	Bidimensional datasets (Figures 8-10) showed a very intimate connection between accuracy and AMI, as was expected due to the clustering hypothesis of SSL. On higher dimensional datasets (Figures 11-16), however, the connection between these metrics is not as intimate, as it was observed in Digit1 (\autoref{fig11}), Ringnorm (\autoref{fig13}), Texture (\autoref{fig15}) and USPS (\autoref{fig16}). The exact reason for this behavior is unclear to us and demands a deeper investigation.
	\par 
	Now, regarding our tuning method, we see that setting $\gamma=\frac{1+q}{2q}$ can lead to a faster execution time than $\gamma = 1$ as one would expect from our study in the previous section since the first approach produces lower values of $\beta^{*}_{\gamma}$ (\autoref{fig7}). This behavior is observed in a more pronounced way in Two moons (\autoref{fig8}), Digit1 (\autoref{fig11}), Twonorm (\autoref{fig12}) and Ringnorm (\autoref{fig13}). In other datasets, both methods have a similar execution time, which leads us to believe both values of $\beta^{*}_{\gamma}$ fall into a plateau of bad computational performance.  
	\par 
	The former behavior may be related to the limitations of the approximation constructed in the previous chapter, since higher values of $q$ lead to higher values of $\beta^{*}_{\gamma}$ (\autoref{fig7}). In fact, for bidimensional datasets, which have $q\leq 5$, NMF shows its better performance. In the high-dimensional case, NMF with $\gamma=\frac{1+q}{2q}$ outperforms other approaches consistently when $q=2$ (Figures 11-13). As the number of classes increases, values of $\beta^{*}_{\gamma}$ fall in the region of worst performance (Figures 14-16) and LGC becomes the fastest algorithm. Also, results presented in \autoref{sec:5} suggest that this slowing down cannot be overcome by addition of labeled data, as most datasets tend to become slower with the increase in $r_{l}$. 
	\par
	When we analyze results on accuracy and AMI as functions of $r_{l}$ we observe all algorithms have a similar behavior of improving those metrics with the addition of labeled data. They diverge, however, on how susceptible they are to this increase, as can be seen in sets like Ringnorm (\autoref{fig13}) and Landsat (\autoref{fig14}). 
	\par 
	Looking at the $k-$dependency of accuracy and AMI there is a well-defined tendency of increasing these quantities for lower $k$, which can then be followed by a decrease that is much less sensitive to variations in the number of nearest neighbors, as is the case of Digit1 (\autoref{fig11}). On Texture (\autoref{fig15}) and USPS (\autoref{fig16}), however, we see that different algorithms respond differently to different topologies of $G_{D}$, with Potts-based methods like GRF and NMF being more tolerant to changes in $k$ in terms of accuracy and AMI. 
	\par 
	It is also noteworthy that setting $\gamma=\frac{1+q}{2q}$ for NMF showed a more stable profile of accuracy and AMI as a function of $k$ than other methods. This can be a property of interest, as choosing $k$ for an application is a hard problem overall due to the aforementioned behavior of other algorithms.
	
\section{Conclusion}

	We have studied the problem of tuning $\beta$ in the NMF equations for the Potts model in applications of semi-supervised transductive classification. Through an analysis of different quantities related to the problem and the model, we were able to verify the difficulty of the problem, as optimal results are usually associated with higher computational times. As labeled data becomes scarcer, finding the best classifications can become even harder due to a well-pronounced peak in quantities like accuracy and AMI.
	\par 
	By the analysis of the probability of the most probable configuration, however, we were able to identify more stable classifications with higher probabilities. By using an approximation for $\Gamma$ we then tested two tuning methods by using two different target values $\gamma=\frac{1+q}{2q}$ and $\gamma=1$ for the said quantity. Results then showed that proposed approaches are effective and can improve on classical algorithms like GRF and LGC, particularly on datasets with fewer classes.
	\par 
	Our studies also raises questions to the possibility of achieving better computational performance in datasets with higher $q$ while maintaining the good metrics on accuracy and AMI. This of course demands novel tuning methods to be proposed as well as the study of novel models for the task at hand.
	\par 
	Regarding the area of SSL, our most interesting contribution was the observation that setting $\gamma=\frac{1+q}{2q}$ leads to a more smooth dependency of accuracy and AMI as a function of $k$ when compared to the other methods, which might be of interest to practitioners in this field.
	\par 
	Overall, our work helps illustrate the problem of parameter tuning in the studied model for SSL. We hope our efforts draw the attention of more researchers to this interesting problem, as this paper in no way can claim it has conquered it. We want, however, to keep our eyes on it and elaborate novel approaches as well as study other approximations and models in order to advance our understanding.

\section*{Acknowledgments}

This study was financed in part by the Coordenação de Aperfeiçoamento de Pessoal de Nível Superior – Brasil (CAPES) – Finance Code 001. We also thank professsors Denis Salvadeo from IGCE/UNESP, Alexandre Levada from DC/UFSCar and the anonymous reviewer of a previous draft of this work for insightful comments and discussions. \\

\bibliographystyle{unsrt} % We choose the "plain" reference style
\bibliography{biblio.bib}

\end{document}